\documentclass{article} %
\usepackage{latex/iclr2026_conference,times}

\usepackage{amsmath,amsfonts,bm}

\def\eqref#1{equation~\ref{#1}}

\def\1{\bm{1}}

\DeclareMathAlphabet{\mathsfit}{\encodingdefault}{\sfdefault}{m}{sl}
\SetMathAlphabet{\mathsfit}{bold}{\encodingdefault}{\sfdefault}{bx}{n}

\usepackage[pagebackref=true,breaklinks=true,colorlinks,citecolor=gray]{hyperref}
\usepackage{url}
\usepackage{comment}
\usepackage{algorithm}
\usepackage{subcaption}
\usepackage{algpseudocode}
\usepackage{graphicx}
\usepackage{booktabs}
\usepackage{multirow}
\usepackage{placeins}

\makeatletter
\def\thanks#1{\protected@xdef\@thanks{\@thanks
        \protect\footnotetext{#1}}}
\makeatother

\title{Self-Improving Loops\\ for Visual Robotic Planning}

\author{Calvin Luo\textsuperscript{*,1}
\thanks{*: Equal contribution. Correspondence to: \href{mailto:calvin_luo@brown.edu}{calvin\_luo@brown.edu} and
\href{mailto:zilai_zeng@brown.edu}{zilai\_zeng@brown.edu}.},
Zilai Zeng\textsuperscript{*,1}, 
Mingxi Jia\textsuperscript{1}, 
Yilun Du\textsuperscript{2}, 
Chen Sun\textsuperscript{1}  \\
\textsuperscript{1}Brown University, 
\textsuperscript{2}Harvard University}

\iclrfinalcopy %

\begin{document}

\maketitle

\vspace{-0.65em}
\begin{abstract}
\vspace{-0.25em}
Video generative models trained on expert demonstrations have been utilized as performant text-conditioned visual planners for solving robotic tasks. However, generalization to unseen tasks remains a challenge. Whereas improved generalization may be facilitated by leveraging learned prior knowledge from additional pre-collected offline data sources, such as web-scale video datasets, in the era of experience we aim to design agents that can continuously improve in an online manner from self-collected behaviors.  In this work we thus propose the Self-Improving Loops for Visual Robotic Planning (SILVR), where an in-domain video model iteratively updates itself on self-produced trajectories, and steadily improves its performance for a specified task of interest.  We apply SILVR to a diverse suite of MetaWorld tasks, as well as two manipulation tasks on a real robot arm, and find that performance improvements continuously emerge over multiple iterations for novel tasks unseen during initial in-domain video model training.  We demonstrate that SILVR is robust in the absence of human-provided ground-truth reward functions or expert-quality demonstrations, and is preferable to alternate approaches that utilize online experience in terms of performance and sample efficiency. Visualizations and code are provided at \href{https://diffusion-supervision.github.io/silvr/}{\nolinkurl{diffusion-supervision.github.io/silvr/}}.
\end{abstract}

\vspace{-0.75em}
\section{Introduction}
\vspace{-0.15em}
Advancements in video generative modeling capabilities have directly led to their increased utilization as visual planners for robotic applications~\citep{du2024video, yang2023learning,  ko2024avdc, liang2024dreamitate}. The synthesized visual plan, in the form of video frames generated with text conditioning, can be translated into executable actions via inverse dynamics models (IDMs). Intuitively, the data on which the video generative models and the IDMs are trained can greatly impact robotic performance and generalization.  When explicitly optimized on in-domain examples of expert behavior, such visual planners are able to synthesize successful plans for solving demonstrated tasks in a robust manner.  However, for arbitrary robotic behaviors, expert-quality demonstrations may not be readily available, and collection may be prohibitively expensive. It is thus worthwhile to investigate how visual planning can automatically \textit{adapt} and \textit{generalize} to novel tasks of interest.

Recent work has investigated how base generalization performance for visual planning can be improved by integrating knowledge from large-scale datasets of text and video collected from the internet.  Adapt2Act~\citep{luo2025solving} creates a powerful, generalizable, text-conditioned visual planner by combining a large-scale model pretrained on web-scale video data with a video model trained on a small set of in-domain demonstrations via score composition.  At a high level, the adapted video model draws upon large-scale motion priors and powerful zero-shot text conditioning capabilities from the web-pretrained video model to facilitate generalization. Simultaneously, it can leverage the in-domain video model to better generate visual plans that respect the environment-specific visual characteristics and dynamics of the robotic setting.  The result is an adapted video model that can generate in-domain-appearing plans for novel, unseen tasks conditioned on natural language.

\begin{figure}[t]
    \centering
    \includegraphics[width=\linewidth]{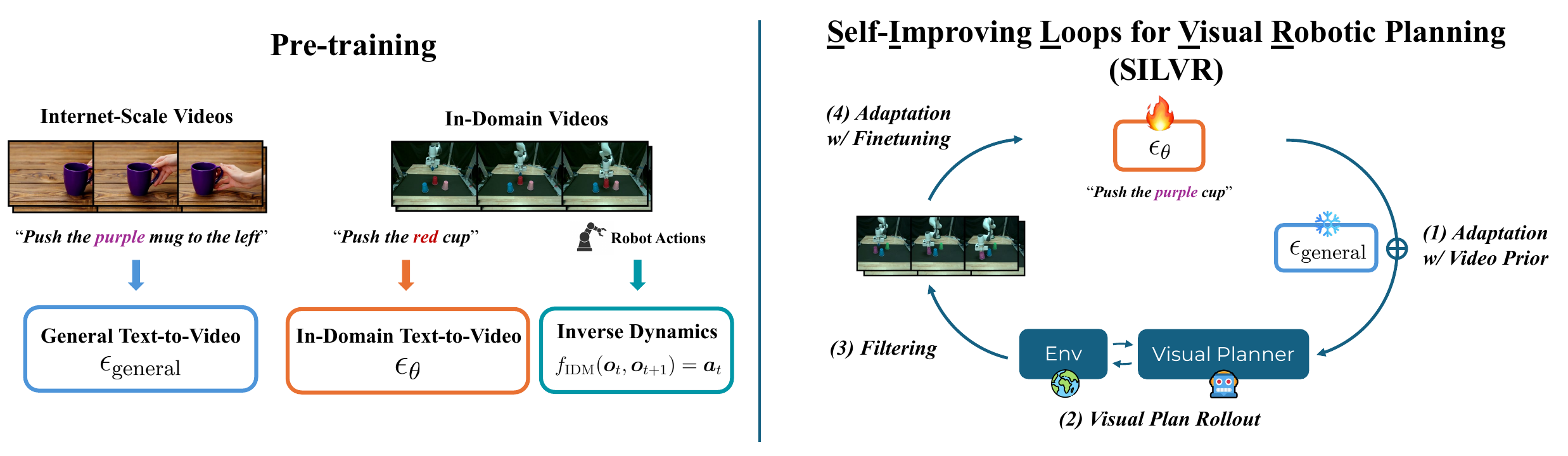}
    \caption{\textbf{SILVR Framework.}
    SILVR has access to two pretrained video generative models (left): one pretrained generally on internet-scale data and another pretrained on a general set of in-domain demonstrations.  By default, SILVR uses the in-domain video model as a visual planner, which when utilized to interact with the environment, is able to achieve successful trajectories even for initially unseen tasks.  These trajectories are then iteratively fed back to finetune the in-domain model (right), thus improving the overall quality of future visual planning as a whole through self-collected online experience.  SILVR can optionally incorporate internet-scale pretrained video models as prior, which particularly improves performance in the case of real-world robotic experiments.}
    \label{fig:sail_teaser}
    \vspace{-5pt}
\end{figure}

Despite extending the amount of data utilized for visual planning to internet-level, the model still only has access to purely offline data. In the era of experience, we aim to design agents that can continuously improve from self-collected behaviors and feedback. In such a way, the agent can break free beyond the limits of offline data and learn by itself to refine performance on a specified task of interest.  We therefore propose \textbf{\underline{S}elf-\underline{I}mproving \underline{L}oops for \underline{V}isual \underline{R}obotic Planning (SILVR)}, where the generated video iteratively self-improves with online experience, particularly with respect to behaviors previously unseen in the initial dataset of environment demonstrations. As shown in Figure~\ref{fig:sail_teaser}, each loop adapts the video generative model (and optionally, the IDM) with environment-grounded data collected by the robotic agents following their own-generated visual plans.

SILVR utilizes a sparse reward signal to filter online experience for further finetuning of the visual planner; however, it is quite natural to consider alternative methods beyond visual planning or direct finetuning of the generative model.  In our experiments we demonstrate that visual planning is superior to direct behavior cloning in both initial generalization performance as well as self-improvement capability, given the same amount of offline data and online experience.  Furthermore, the final resulting visual planner can be distilled into a lightweight policy via behavior cloning for both fast and performant decision-making.  We further showcase how SILVR is more sample efficient than reinforcement learning finetuning approaches, making it more applicable to real-world robotic settings.  SILVR is also robust to the quality of the sparse reward signal; rather than requiring a human-defined ground-truth success function, we demonstrate that iterative improvements still arise when utilizing a pretrained vision-language model (VLM) to score experience based on the task descriptions.

To summarize, SILVR enables superior sample-efficient self-improvement over initially unseen tasks through visual planning, naturally integrating internet-scale pretrained video priors over text-alignment and motion when necessary, in comparison to regular action-prediction policies. %
During deployment, a policy can be ``distilled'' from SILVR's components for fast inference, showcasing how SILVR facilitates effective generalization and self-improvement for robotic tasks without sacrificing final execution speed.

We perform extensive evaluations of SILVR on the MetaWorld task suite, focusing on novel tasks unseen during initial training of the in-domain video model. We discover that the success rate of following visual plans synthesized through SILVR indeed continuously improves, by as much as \textbf{285\%} over 10 iterations.  We also apply SILVR to a real-world robot arm for two distinct manipulation tasks: selecting and pushing a colored object, and selecting and opening a colored drawer.  We show how SILVR also naturally enables the incorporation of priors from internet-scale pretrained video generative models to facilitate task and visual generalization in real-world visual settings and dynamics.  We demonstrate that performance for color combinations unseen during the initial offline training improves over multiple iterations through SILVR. 

\section{Related Work}

\textbf{Video Generation for Decision Making.} Recent advances in video models have achieved unprecedented visual quality and physical fidelity for video synthesis~\citep{guo2023animatediff, yang2024cogvideox, videoworldsimulators2024, veo2, wan2025}. This has demonstrated promise in summarizing world dynamics through videos~\citep{yang2024video, Bruce2024GenieGI} and has inspired the application of video models to solving decision-making problems~\citep{alejandro2023viper, du2024learning, yang2023learning, McCarthy2024TowardsGR, liang2024dreamitate}. Prior works have utilized video generative models as reward functions~\citep{luo2024text, alejandro2023viper, huang2023diffusion}, dynamics models~\citep{yang2023learning, Bruce2024GenieGI, Valevski2024DiffusionMA}, and pixel-based planners~\citep{ko2024avdc, ajaj2023hip, du2024video, Zhou2024RoboDreamerLC}. As in UniPi~\citep{du2024video}, we employ video models to predict text-conditioned visual plans that depict future outcomes, which are subsequently translated into actions via inverse dynamics. While the performance of such visual planners may often be limited by their offline pretraining data, our approach allows iterative improvement by learning from online environment interactions.

\textbf{Self-Improving Generative Models.} Continuously improving by learning from self-produced cumulative experience is an essential capability of intelligent agents. Prior work has demonstrated the effectiveness of improving LLMs with their self-generated outputs~\citep{yu-etal-2024-teaching, Tian2024NEURIPS_Toward_Self_Improvement, Huang2022CONFERENCE_Large_Language_Models}, where the LLM can serve as its own reward function~\citep{Yuan2024ICML_Self_Rewarding_Language} for preference optimization or data synthesizer~\citep {Patel2024ARXIV_Large_Language_Models} for supervised finetuning. However, a similar self-improvement recipe for video generation models remains underexplored. Most relevant to our work, VideoAgent~\citep{Soni2024VideoAgentSV} refines video generation through self-conditioning consistency and feedback from a VLM, and collects the successful plan rollouts for finetuning. We instead base our improvement loop on self-adaptation, where we leverage internet-scale video priors to synthesize improved visual plans for tasks unseen during initial in-domain training. Furthermore, our approach can still achieve self-improvement even with an initial model trained on suboptimal data and a notable relaxation on filtering requirements for finetuning data.

\textbf{Reinforcement Learning Finetuning of Behavior Cloning Policies.}
Behavior cloning approaches such as Diffusion Policy~\citep{chi2023diffusionpolicy}, which implements a policy as a diffusion model trained on offline-collected experience, are a performant approach for decision-making.  There have been numerous approaches for finetuning pretrained diffusion policies with respect to online experience and rewards.  ResIP~\citep{ankile2025imitation} utilizes a frozen diffusion policy model to propose action predictions, and learns a policy on top using reinforcement learning that transforms it into a more accurate action to perform in the environment.  DPPO~\citep{ren2024diffusion} treats the sampling procedure of a diffusion policy as an internal Markov Decision Process, and explicitly finetunes the weights of the pretrained diffusion policy with respect to achieved rewards.  DSRL~\citep{wagenmaker2025steering} utilizes a frozen diffusion policy in a deterministic fashion to map noise samples to action samples, and learns a noise selector through reinforcement learning.  In this work, we show that SILVR is more sample efficient than reinforcement learning finetuning of behavior cloning policies, and can achieve faster iterative improvements with respect to online experience.

\vspace{-0.5em}
\section{Method}
\vspace{-0.25em}

We introduce the Self-Improving Loop for Visual Robotic Planning (SILVR), in which a video generative model initially trained on a general set of in-domain demonstrations iteratively improves its visual planning performance for a particular task of interest in a self-adaptive manner. In Section~\ref{sec:method_probadap}, we describe how a small in-domain video model can be integrated with a generally pretrained text-to-video model to produce a strong, generalizable in-domain visual planner for real-world visual settings.  Finally, in Section~\ref{sec:sail}, we demonstrate how SILVR bootstraps an in-domain video model into a high-performing visual planner for solving a novel robotic control task through iteratively fine-tuning on self-collected experience.

\subsection{Video Models as Visual Planners}
\label{sec:method_visual_planner}
Synthesizing a visual plan in imagination and then executing it by converting it into actions is an intuitive and effective way to utilize video generative models for decision making.  Prior work has applied text-guided video generation successfully for task planning~\citep{du2024learning, du2024video, ajaj2023hip, luo2025solving}, across a variety of robot configurations and environment settings.

Specifically, we base our implementation on the UniPi framework~\citep{du2024learning}, in which a text-to-video model is used to synthesize a text-conditioned sequence of future frames as a task plan.  To physically realize the plan, we use a separately trained inverse dynamics model (IDM) to translate consecutive pairs of visual frames into executable robotic actions, which are then directly performed in interaction with the environment.  Visual planning offers the practitioner flexible computational tradeoffs; at a high level, replanning often incurs high computational cost but generally increases accurate plan following, whereas replanning infrequently is cheap but may suffer from error compounding.
In this work, we focus on how such a video generative model can generalize and self-adapt to a novel task of interest through online self-collected experience.

\begin{figure}[t]
\begin{minipage}{\linewidth}
\begin{algorithm}[H]
    \caption{Self-Improving Loops for Visual Robotic Planning (SILVR)}
    \label{algo:sail_pseudocode}
    \hspace*{\algorithmicindent} \textbf{Input:} Initial in-domain video model $\epsilon_{\theta}$, Inverse dynamics model $f_{\text{IDM}}$, Frozen internet-pretrained video prior $\epsilon_{\text{general}}$, Number of iterations $K$, Number of rollouts per iteration $N$, Environment $\texttt{env}$, Task prompt $g$, In-domain initial training data $\mathcal{D}_{\text{ini}}$, Data filter $f_{r}$ \\
    \hspace*{\algorithmicindent} \textbf{Output:} Self-improved in-domain video model $\hat{\epsilon}_{\theta}$, Optional distilled policy $\pi$
    \begin{algorithmic}[1]
        \State $\hat{\epsilon}_{\theta} \gets \epsilon_{\theta}$
        \State $\mathcal{D} \gets \mathcal{D}_{\text{ini}}$ or $\phi$\Comment{Initialize finetuning data with $\mathcal{D}_{\text{ini}}$ or an empty set}
        \For{$i = 1,...,K$}
        \State $\mathcal{D}_{\text{self}} \gets \phi$
        \State $\tilde{\epsilon} \gets \texttt{Adaptation}(\hat{\epsilon}_{\theta}, \epsilon_{\text{general}}, g)$
            \For{$j = 1,..., N$}
                \State $\texttt{env.reset}(g)$
                \State $\mathcal{D}_{\text{self}} \gets$ $\mathcal{D}_{\text{self}}$ $\cup$ $f_{r}(\texttt{Visual\_Planning\_Rollout}(\texttt{env}$, $\tilde{\epsilon}$, $f_{\text{IDM}}$)) %
            \EndFor
        \State $\mathcal{D} \gets \mathcal{D} \cup \mathcal{D}_{\text{self}}$ 
        \State Finetune in-domain model $\hat{\epsilon}_{\theta}$ with accumulated data $\mathcal{D}$\Comment{$f_{\text{IDM}}$ can be optionally finetuned}
        \EndFor
        \State Optionally distill $\hat{\epsilon}_{\theta}$ into a lightweight policy $\pi$
        \State \textbf{return} $\hat{\epsilon}_{\theta}$, $\pi$
    \end{algorithmic}
\end{algorithm}
\end{minipage}
\end{figure}

\subsection{Inverse Probabilistic Adaptation}
\label{sec:method_probadap}
Prior work~\citep{luo2025solving} has investigated how in-domain demonstration data can best be integrated with large-scale pretrained video models for generalizable visual planning; in this work we leverage similar insights to successfully integrate on-the-fly experience into visual planners for iterative self-improvement.  Inverse Probabilistic Adaptation~\citep{luo2025solving, yang2023probabilistic} (IPA) is a training-free approach that adapts generally pretrained text-to-video models for domain-specific video generation.  To perform adaptation, the score predicted by an in-domain video model $\epsilon_{\theta}$ trained on a small sample of demonstrations is composed with the score prediction of a web-scale pretrained model $\epsilon_{\text{general}}$ during the sampling procedure, as depicted in the function below:
\begin{align}
    \tilde{\epsilon}_{\text{inv}} = \epsilon_{\text{general}}(\tau_t, t) + \alpha\Big(\epsilon_{\text{general}}(\tau_t, t \mid \text{text}) + \gamma \epsilon_{\theta}(\tau_t, t \mid \text{text}) - \epsilon_{\text{general}}(\tau_t, t)\Big)
    \label{eq:inv_probadap}
\end{align}
where $\gamma$ is the prior strength, and $\alpha$ is the guidance scale of text-conditioning.  Intuitively, the small in-domain text-to-video model serves as a probabilistic knowledge prior that guides the generation process of the small in-domain model during sampling.  Prior work~\citep{luo2025solving} has found that a visual planner constructed through IPA exhibits both strong generalization capability and in-domain understanding; it is able to synthesize performant visual plans that appear in-domain even for novel tasks unseen during video model training.  This may stem from the fact that IPA utilizes the large-scale pretrained model, which inherently has stronger text-conditioned generalization, as the main denoiser. While \citet{luo2025solving} based their conclusions on experiments in simulated environments, we believe the true promise of web-scale pretrained models lies in their powerful prior for real-world generalization scenarios, as demonstrated by our Panda arm object manipulation evaluations.

\subsection{Self-Improving Loops for Visual Robotic Planning}
\label{sec:sail}

For visual planning approaches, task performance is fundamentally a fixed function of the video models used, and by extension, the data observed.  Even when utilizing IPA, which can improve text-conditioned generalization to novel tasks by effectively increasing the amount of offline data utilized to internet-scale, performance is set after adaptation.  As a result, in this paper, we wish to design agents that can not only leverage offline data as a helpful prior for generalization, but also extend beyond it to continuously improve from self-collected online experience data.

We therefore propose \textbf{Self-Improving Loops for Visual Robotic Planning (SILVR)}, a framework that combines offline data with online experience to create a visual planner that iteratively improves for a particular task of interest.  SILVR is initialized with an in-domain video model $\epsilon_{\theta}$ pre-trained on a set of task demonstrations within the environment.  In each iteration, the in-domain video model is optionally integrated with a large-scale pretrained video model $\epsilon_{\text{general}}$ through IPA.  The video model is utilized as a visual planner to interact with the environment and solve tasks not necessarily observed in the initial training stage; in SILVR, the trajectories collected through this interaction are used for further finetuning of the in-domain video model (as shown in Algorithm~\ref{algo:sail_pseudocode}).  As the in-domain model adapts to its own self-collected experience from deployment on a novel task, it improves its ability to solve that particular task over time.  In this way, SILVR iteratively bootstraps an in-domain video model into a strong visual planner for a novel task of interest through a self-adapting improvement cycle.

We demonstrate that the visual planning approach enables strong self-improvement in a virtuous loop in a sample efficient manner, compared to reinforcement learning finetuning of behavior cloning models.  We further stress-test our framework through ablations, and demonstrate that ground-truth human-defined reward functions can be replaced with an automated VLM success signal without inhibiting iterative self-improvement from occurring, and that SILVR can handle suboptimal initial data quality
We find that SILVR is a robust approach for iteratively adapting to a task through effective utilization of both offline data and online experience, and reduces requirements on human-supplied components both in terms of feedback and demonstration quality.

Whereas visual planning demonstrates strong self-improvement capabilities, and can flexibly integrate in benefits from large-scale pretrained video models, it can be slow in execution compared to direct policy approaches.  However, after applying SILVR, the final video planning components can be distilled into a lightweight policy through behavior cloning for future inference.  We demonstrate in our experiments that such a final distilled policy achieves higher performance than applying self-improvement techniques to such a policy from the start, and can even demonstrate slight improvement over the final performance of the visual planner teacher.  Thus, SILVR enables a successful balance between visual planning for improved and sample-efficient utilization of online experience, and a lightweight distilled policy for reactive inference during deployment.

\section{Experiments}

We investigate how SILVR can improve an in-domain video model initially trained on a limited set of demonstrations and tasks to further solve novel robotic control tasks through self-collected experience.  We focus on two main robot settings to evaluate SILVR: the MetaWorld-v2~\citep{yu2020meta} simulated environment, and a real-world Franka Emika Panda robot arm.  We describe our experimental setup for each environment, as well as different design decisions considered.

\subsection{Experimental Setup and Evaluation}
\label{sec:experimental_setup}
\textbf{Synthetic Environment:} MetaWorld encompasses a wide selection of tasks, allowing us to thoroughly assess visual planning performance trends through SILVR for many choices of held-out novel tasks.  Furthermore, MetaWorld provides ground-truth success evaluations, enabling strictly quantitative comparisons on task performance as well as iterative improvement abilities.  For MetaWorld experiments, we first collect 25 demonstrations from 8 different tasks (denoted with an asterisk in Table~\ref{table:text_prompts}) for initial in-domain video model and inverse dynamics model training. Subsequently, we instantiate SILVR with both models on 12 unseen tasks for self-improvement. In each SILVR iteration, we collect 30 trajectories rendered from the environment via visual planning, and finetune both in-domain video model and inverse dynamics model using filtered successful data. Due to the sim-to-real gap, we disable the use of internet-scale video prior in MetaWorld by default.

\begin{table}[t]
\centering
\setlength{\tabcolsep}{4.8pt}
\scalebox{0.9}{
\begin{tabular}{@{}lccccc@{}}
\toprule
Iteration & 0               & 1               & 2               & 3               & 4                \\ \midrule
DSRL (w/ GT Filter)      & 9.4 ($\pm$1.7) & 8.3 ($\pm$1.6)  & 7.4 ($\pm$0.9)  & 7.5 ($\pm$3.4)  & 7.7 ($\pm$3.4)   \\
BCIL (w/ GT Filter)      & 5.6 ($\pm$0.6)  & 12.3 ($\pm$2.3) & 20.9 ($\pm$1.6) & 23.3 ($\pm$0.4) & 23.2 ($\pm$0.9)  \\
SILVR (w/ GT Filter)     & 14.7 ($\pm$0.6) & 27.7 ($\pm$1.9) & 33.5 ($\pm$2.2) & 43.5 ($\pm$2.6) & 44.2 ($\pm$4.5)  \\
SILVR (w/ VLM Filter)      & 17.0 ($\pm$0.6) & 24.4 ($\pm$0.9) & 28.7 ($\pm$2.8) & 34.4 ($\pm$1.4) & 38.4 ($\pm$1.3)  \\\midrule
SILVR-Distilled DP &                 &                 &                 &                 & 44.2 ($\pm$4.5) $\rightarrow$ 49.2 ($\pm$3.4) \\ \bottomrule
\end{tabular}
}
\caption{\textbf{SILVR Results on MetaWorld}. We report the average performance over 12 unseen MetaWorld tasks for SILVR and all baseline methods, each aggregated over three seeds. We also provide the performance of diffusion policy distilled from the video model from the last SILVR iteration, denoted as ``SILVR-Distilled DP". SILVR outperforms all baselines by a large margin since Iteration 1. Furthermore, SILVR-Distilled DP achieves the best overall performance.}
\label{table:silvr_baseline_results}
\end{table}

\begin{figure}[t]
    \centering
    \includegraphics[width=\linewidth]{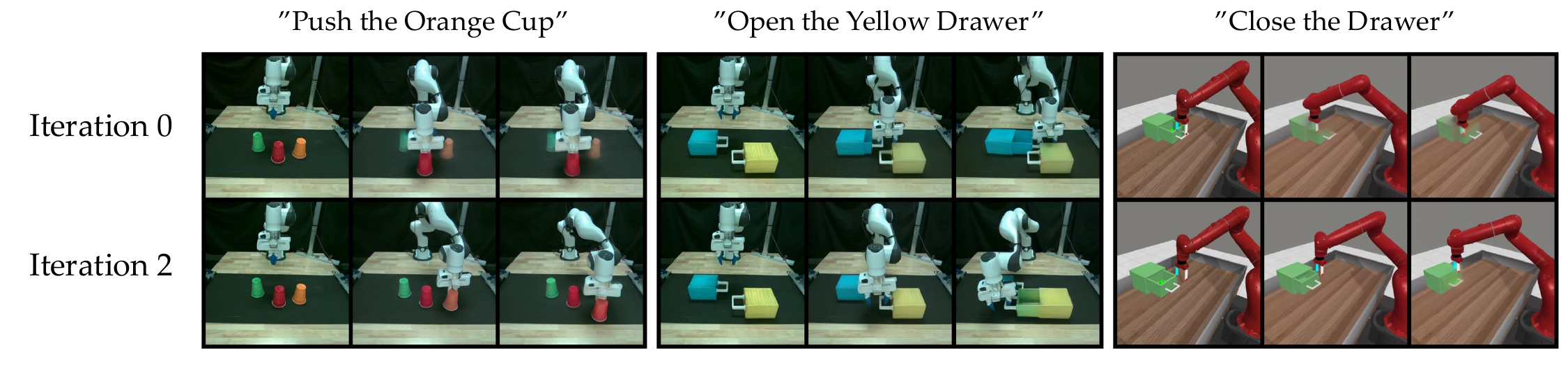}
    \caption{\textbf{Qualitative results on visual plans improvement.} We illustrate visual plans for a variety of tasks and settings at Iteration 0 (top) and Iteration 2 (bottom) with random initial object locations. Although the visual plan at Iteration 0 renders blurry objects and fails to complete the specified tasks, our approach synthesizes the correct visual plan (with slight color drift) after two SILVR iterations.}
    \label{fig:franka_sail_visual_plans_comparison_all_tasks}
\end{figure}

\textbf{Real-World Environment:} Deploying SILVR on a robot arm in the real world demonstrates the practicality of the approach, as well as tests robustness to real-world confounding factors such as lighting conditions.  In one experiment, we utilize a Franka Emika Panda robot arm for the task of pushing cups specified by a user-provided text prompt.  In contrast to the MetaWorld setups, where each task of interest has its own distinct visual setting, we construct the cup experiment as a consistent scene setting of 3 differently colored cups (Figure~\ref{fig:sail_teaser}). Success is then measured in terms of whether the robot arm can accurately locate a specified color cup and push it forward.  To test generalization, conditioned on natural language, we evaluate successful planning and execution performance on unseen cup colors.  In practice, we use a set of four colors (red, green, blue, pink) for in-domain training and two novel colors for testing generalization (orange, purple).  This translates to 12 possible unique tasks formed from combinations of the seen colors, and we train our in-domain video model with 10 human-teleoperated demonstrations of each for a total of 120 training videos. Then, generalization evaluation is calculated as an average over 5 rollouts for every possible pair combination of the seen color set combined with the novel color, for a total of 30 videos. For both novel colors, we initialize SILVR using the same pretrained in-domain video model. In each SILVR iteration, we combine self-collected data with the initial demonstrations for in-domain finetuning.

In a second real-robot experiment, we utilize the Panda arm to select and open a drawer specified via a user-provided text prompt.  The scene is constructed as two distinctly colored closed drawers, where the robot is prompted with one particular color and expected to open its corresponding drawer.  We use a set of three colors (red, green, blue) for in-domain training and one novel color (yellow) for testing generalization.  With 24 possible drawer placement combinations for each ordered pair of seen colors, of which there are six, this amounts to a total of 144 human-teleoperated demonstration training videos.  Consistent with the cup pushing experiment, we use half the possible combinations for evaluation; therefore, performance is calculated as an average over 12 rollouts for every possible pairing of the novel color with a seen color, for a total of 36 self-collected trajectories per iteration.  

For both real-robot experiments, success is judged by a human for evaluation. The same success signal is also used to perform data filtering on the rollouts. We study the impact of data filtering in Section~\ref{sec:filtering_signal_silvr}, and enable adaptation with internet video priors through IPA by default.

\begin{figure}[t]
    \centering
    \includegraphics[width=\linewidth]{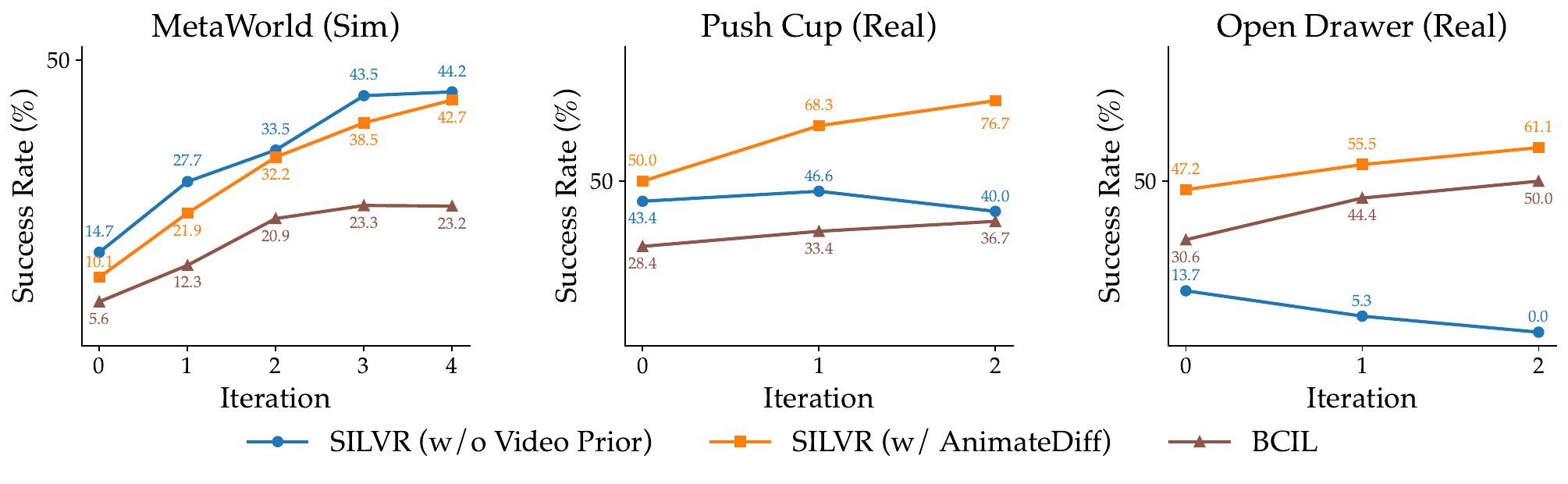}
    \caption{\textbf{SILVR Results} in comparison to Behavior Cloning Improvement Loop (BCIL). We report the average performance over 12 unseen MetaWorld tasks, as well as novel pushing and drawer opening tasks for Panda arm experiments across several iterations of self-improvement (x-axis). Numbers in the graph correspond to success rate achieved (y-axis).}
    \label{fig:silvr_mw_franka_main}
\end{figure}

\textbf{Implementation Details:} We implement our in-domain video model based on AVDC~\citep{ko2024avdc}, with an added cross-attention layer to each level of the denoising U-Net to further improve text-conditioning capabilities. We train in-domain video models to predict 8 future frames conditioned on the current observation and task prompt, with a frame skip of 1 for MetaWorld and 16 for real-robot experiments. For the large-scale pretrained text-to-video model, we use AnimateDiff~\citep{guo2023animatediff} ($\sim$2B parameters), which is pretrained on WebVid-10M~\citep{bain2021webvid}.  %
Each iteration of SILVR finetunes the in-domain video model for 10,000 steps with a learning rate of 1e-6 on MetaWorld and 1e-5 on Panda Arm drawer opening tasks, and 8,000 steps with a learning rate of 2e-5 on Panda Arm pushing tasks.  We investigate two IDM implementations; one that follows prior work~\citep{luo2025solving} that implements the IDM as a MLP network that takes in the outputs of a VC-1~\citep{majumdar2023vc1} encoder model, finetuned on in-domain demonstrations.  Whereas the MLP-IDM was sufficient for real-world experiments, we found improved performance when using a Diffusion-IDM (DIDM), which is implemented as a Diffusion Policy~\citep{chi2023diffusionpolicy} with an additional goal frame provided as conditioning, for simulated settings.  Additional details on IDM design and hyperparameters are provided in Section~\ref{sec:implement_appendix}.  For MetaWorld experiments, the DIDM is iteratively finetuned on the online collected experience along with the video model.

\begin{figure}[t]
    \centering
    \includegraphics[width=\linewidth]{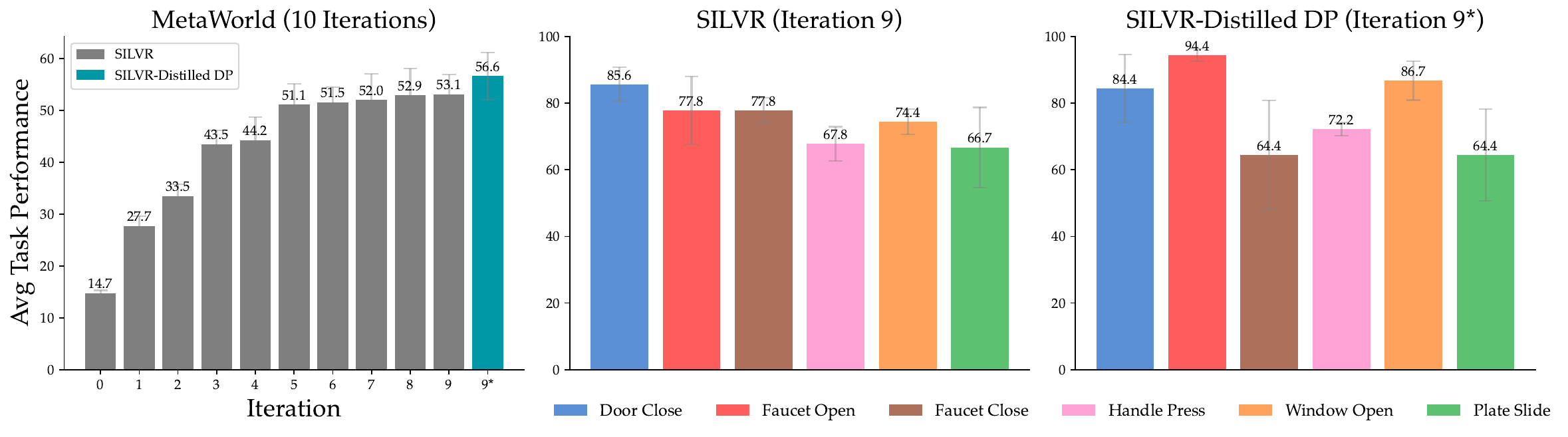}
    \caption{\textbf{SILVR results on MetaWorld for 10 iterations}. We report effects of training SILVR on an extended amount of iterations.  On the left plot, we show that performance continues to monotonically increase, but with diminishing improvements and effective saturation past iteration 5. On the middle and right plots we visualize a comparison between the final iteration visual planner against its distilled student BC policy from the visual planner across 6 tasks, where we observe that certain tasks actually improve after distillation.}
    \label{fig:silvr_iter10}
    \vspace{-1em}
\end{figure}

\subsection{SILVR via Filtered Finetuning}

We report incremental visual planning results for MetaWorld through five SILVR iterations, against two self-improving baseline methods built off of Diffusion Policy (DP)~\citep{chi2023diffusionpolicy}: DSRL~\citep{wagenmaker2025steering} and Behavior Cloning Improvement Loop (denoted as ``BCIL''). We initialize DSRL and BCIL with a Diffusion Policy trained on the same in-domain data as used for in-domain video model and inverse dynamics model pre-training in SILVR. We utilize the ground-truth task success signal to filter the same amount of self-collected data per iteration and finetune models with only successful trajectories.

In Table~\ref{table:silvr_baseline_results}, the performance is averaged over 12 unseen MetaWorld tasks and aggregated over 3 seeds. Despite the same initial in-domain data being shared across all methods, we observe that the visual planning approach achieves better performance than DP-based approaches at Iteration 0, laying solid foundations for subsequent improvement. Compared to DP, which learns to map observations to actions directly, visual planning decouples dynamics modeling from action prediction. We hypothesize that the separately learned environment visual dynamics is easier to transfer when solving a novel task, leading to stronger base generalization performance through visual planning. While DSRL fails to improve and BCIL quickly saturates at a low success rate after several iterations, SILVR continuously improves and consistently outperforms the baselines by a large margin, demonstrating its high sample efficiency.

\begin{figure}[t]
  \centering
  \begin{subfigure}[b]{0.54\textwidth}
    \centering
    \includegraphics[width=\textwidth]{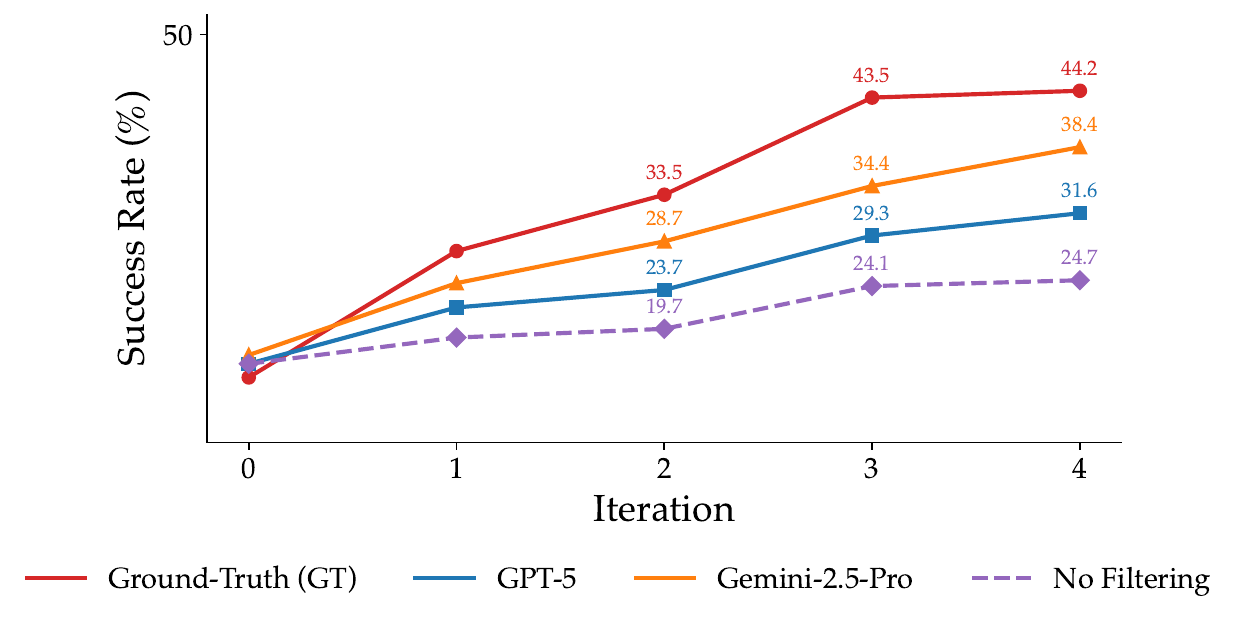}
    \caption{\small MetaWorld (over 12 unseen tasks)}
    \label{fig:sail_mw_filter_ablation}
  \end{subfigure}
  \hfill
  \begin{subfigure}[b]{0.45\textwidth}
    \centering
    \includegraphics[width=\textwidth]{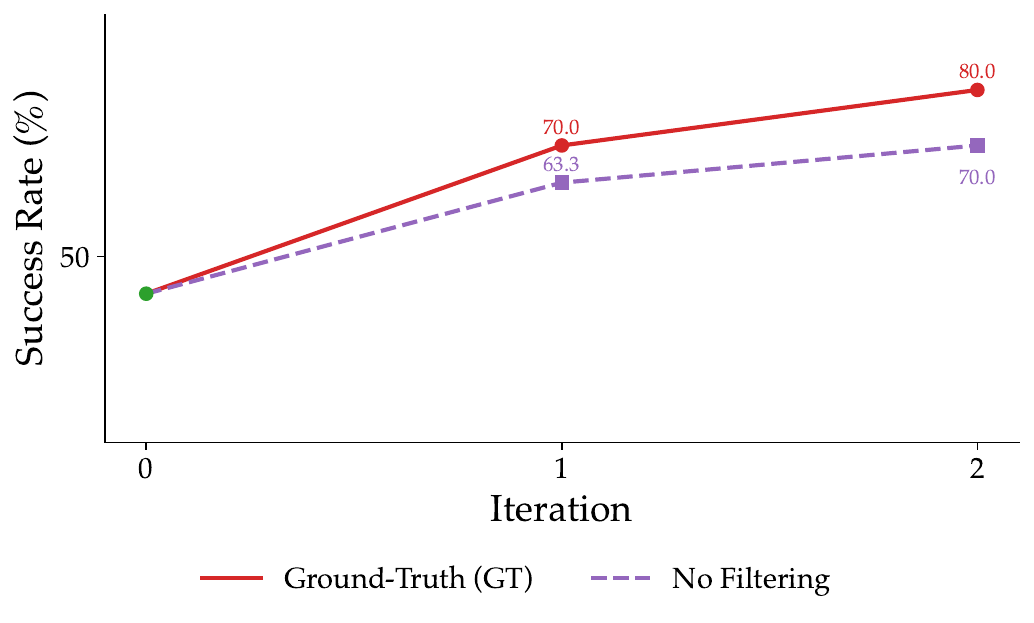}
    \caption{\small Panda Arm Cup Push (Orange)}
    \label{fig:sail_franka_filter_ablation}
  \end{subfigure}

  \caption{
  \textbf{Ablations on data filtering.} We compare the effect filtering has on success rate (y-axis) across iterations of finetuning (x-axis), on both MetaWorld~(\ref{fig:sail_mw_filter_ablation}) and Panda arm~(\ref{fig:sail_franka_filter_ablation}) setups.  On MetaWorld (left plot), we further report accuracy when filtering is performed by a VLM. We observe SILVR consistently improves task even without access to ground-truth filtering signals.}
  \label{fig:main}
  \vspace{-1.5em}
\end{figure}

\vspace{-1.5em}
\subsection{SILVR with Internet-Scale Video Prior}
\label{sec:silvr_with_video_prior}
\vspace{-0.5em}
One important observation from Table~\ref{table:silvr_baseline_results} is that the generalization capability of the visual planner in solving novel tasks has a fundamental impact on self-improving dynamics. When the task of interest is exceptionally challenging or involves confounding factors, the capacity of the in-domain video model alone might not be sufficient to elicit self-improving behaviors. In the real-world experimental setup, we adapt our in-domain model with an internet-pretrained video prior, AnimateDiff, to further strengthen the zero-shot generalization and adaptability of the visual planner. In Figure~\ref{fig:franka_sail_visual_plans_comparison_all_tasks}, we qualitatively illustrate the improvements of generated visual plans after two SILVR iterations in combination with AnimateDiff. Without observing any demonstrations of the specified tasks at Iteration 0, the visual planner can synthesize plans with blurry objects where the robot arm executes the task incorrectly. On the other hand, two iterations of SILVR not only improve the clarity of the visual plans, but also demonstrate successful task completion behaviors in the same initial layout.

In the middle plot of Figure~\ref{fig:silvr_mw_franka_main}, we report the average performance of pushing cups in two unseen colors, orange and purple, aggregated over 30 rollouts per iteration. We discover that SILVR consistently improves over iterations when adapting with AnimateDiff. In the rightmost plot of Figure~\ref{fig:silvr_mw_franka_main}, we provide the SILVR results across iterations on opening a novel colored drawer, averaged over 36 rollouts per iteration, and find that SILVR bootstraps initial visual planning performance with the help of the internet video prior. However, in both real-world experiments, the visual planner struggles to improve or even continuously deteriorates without video prior. This highlights the importance of internet video prior in self-improving behavior under real-world setups with increased visual complexity and task difficulty. In simulated environments like MetaWorld, we observe that the self-improving trend occurs regardless of whether internet video priors are utilized or not, indicating that the benefits of utilizing internet priors may diminish when there is a substantial sim-to-real gap.

Consistent with our findings on MetaWorld, action-predictive behavior cloning has a lower base generalization performance and slower self-improvement trend compared with SILVR on real robots, as shown in the BCIL curves of the middle and right plots in Figure~\ref{fig:silvr_mw_franka_main}.  This highlights a key benefit of SILVR: its ability to seamlessly utilize internet-pretrained video prior for improved text-conditioned generalization and sample-efficient online improvement in real-world robotic settings.

\vspace{-0.5em}
\subsection{SILVR Saturation and Distillation}

To further understand the limits of self-improving behavior over iterations, we provide MetaWorld results for 10 SILVR iterations in Figure~\ref{fig:silvr_iter10}.  We find that SILVR saturates at Iteration 5 with marginal gains in the following iterations.  We hypothesize that this may potentially arise from discovered local minima in task-specific strategy, where similar experiences are collected until saturation.  We believe that a possible mitigation is to introduce the notion of “exploration” into the visual planning framework to avoid ``unimodal'' behavior. Such research may look into how to extract out more diverse plans from the video planner by exploiting the stochastic nature of visual generative models. We leave this investigation as promising future work.

While visual planning approaches demonstrate strong performance in task generalization, their inference speed is bottlenecked by the video generation process, which can be prohibitively expensive for downstream applications. To mitigate this, we distill the video model from the last SILVR iteration into a lightweight diffusion policy. As shown in Figure~\ref{fig:silvr_iter10}, the SILVR-distilled diffusion policy at Iteration 9 significantly outperforms the BCIL baseline and achieves the best overall performance. A slight performance increase after distillation is a trend consistently observed across iterations, such as demonstrated in Iteration 4 of Table~\ref{table:silvr_baseline_results}.  This further demonstrates that SILVR not only excels in sample-efficient task adaptation, but also supports high inference efficiency for downstream deployment via distillation.

\subsection{Impact of Data Filtering Signals on SILVR}
\label{sec:filtering_signal_silvr}
\vspace{-0.5em}
While utilizing self-collected data is a promising approach for scalable self-improvement, filtering collected experience often requires some level of human intervention, whether through manually determining successful trajectories or designing a heuristic for quality control.  We therefore investigate how different filtering techniques affect SILVR performance, or if SILVR is robust to such design decisions.  For both MetaWorld and Panda Arm settings, we compare between using a ground-truth or human-evaluated notion of success to filter what trajectories the in-domain model is finetuned on, against not using any filtering at all and utilizing all achieved trajectories irregardless of outcome. While ground-truth success signals can often be inaccessible, we investigate whether the current state-of-the-art VLMs, GPT-5 and Gemini-2.5-Pro, can provide useful task success signals and serve as a robust alternative to ground-truth signals on MetaWorld.

In Figure~\ref{fig:sail_mw_filter_ablation}, we observe that both GPT-5 and Gemini-2.5-Pro can still enable self-improving behavior across SILVR iterations when serving as a task success judge, in which Gemini achieves the best performance among all VLM filters. We also discover that without any data filtering, the improvement over each SILVR iteration appears to be marginal compared to the filtered setup. On the other hand, in Figure~\ref{fig:sail_franka_filter_ablation}, for the Panda arm, we observe that no filtering still facilitates continuous improvement over every iteration through SILVR when adapted with the internet video prior. This is an encouraging finding, as it suggests that even for settings where manual curation of experience is expensive, self-improvement can still occur.  We attribute this property to score composition~\citep{luo2025solving, yang2023probabilistic}; suboptimal demonstrations can still communicate useful information to the in-domain model, such as visuals, valid motions, and interaction dynamics of the specific deployment environment that when combined with an internet-pretrained video model can result in a final composed output plan can be both performant as well as appearing in-domain.  As such, the success rate may still improve from iteration to iteration without filtering, as the in-domain model improves its modeling of environment dynamics and visuals, even from suboptimal self-collected experience over iterations.

\begin{figure}[t]
    \centering
    \includegraphics[width=\linewidth]{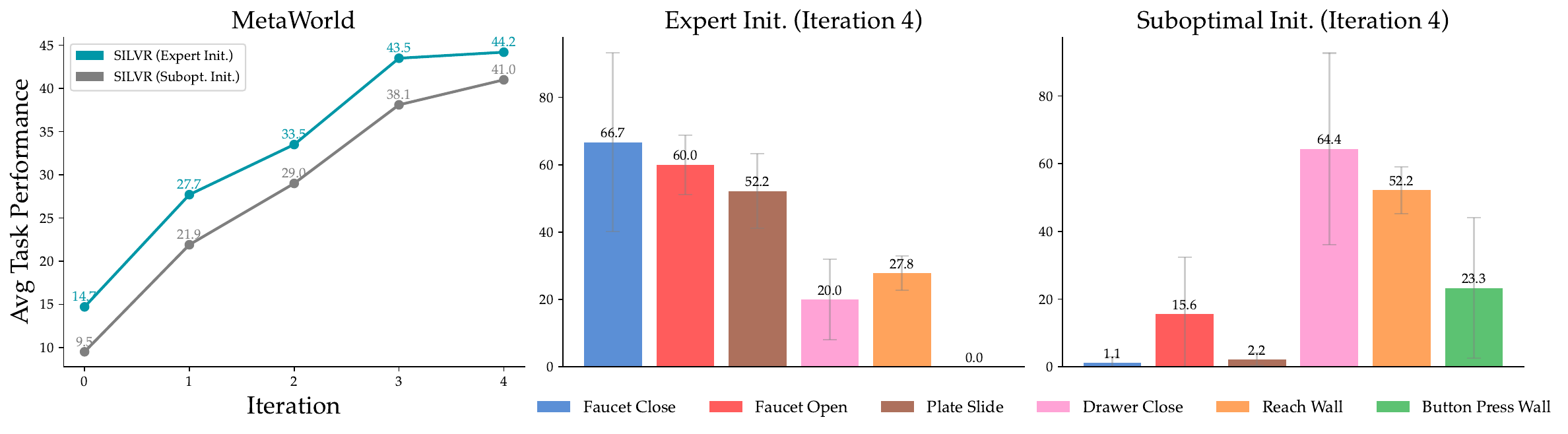}
    \vspace{-20pt}
    \caption{\textbf{Ablation studies on in-domain data quality}. On the left plot, we report SILVR performance averaged over 12 novel tasks and 3 seeds when the provided initial demonstrations of seen tasks are expert-quality or suboptimal.  We find that SILVR successfully self-improves despite suboptimal data initialization.  
    In the middle and rightmost plots, we visualize 6 tasks that have the most distinctive performance between expert and suboptimal SILVR initializations at the final iteration.}
    \label{fig:silvr_expert_vs_suboptimal}
    \vspace{-1.0em}
\end{figure}

\subsection{SILVR with Suboptimal Data}
\label{sec:suboptimal_sail}
\vspace{-0.5em}
Visual planners are usually trained explicitly on expert in-domain demonstrations, which communicate not only environment-specific visual characteristics, physics, and interaction dynamics to the generative model during optimization, but also a notion of success and optimal behavior.  However, for arbitrary environments, such expert-quality in-domain data can be expensive to collect and curate at scale.  On the other hand, suboptimal demonstration data, such as utilizing random actions during the collection procedure, may generally be cheaper to gather; however, training on a large dataset of low-quality data may not result in a performant visual planning model capable of generating plans worth following.  A natural question is how robust SILVR is to initialization data, or whether a performant video planner can still be created when only suboptimal demonstrations are available.

In our setting, we construct suboptimal data as simulated trajectories where 70\% of the time a random action is selected and 30\% an expert action is utilized.  As a consequence of this interaction procedure, the resulting trajectories are unable to successfully solve complex tasks.  
In MetaWorld we find that SILVR still demonstrates continuously improving behavior when initialized with suboptimal data, as shown in the leftmost plot of Figure~\ref{fig:silvr_expert_vs_suboptimal}, highlighting the robustness of SILVR to initial data quality. Additionally, we select 6 tasks whose performance differs most between the expert and suboptimal setup, and report their performance on Iteration 4 in the middle and rightmost plots of Figure~\ref{fig:silvr_expert_vs_suboptimal}. We find that Faucet Open, Faucet Close and Plate Slide benefit most from initialization with expert demonstrations, indicating that the skills acquired from seen tasks can be essentially useful when solving these novel tasks. Meanwhile, Drawer Close, Reach Wall, Button Press Wall can benefit more from random exploration than expert actions from specific seen tasks, leading to stronger performance when the in-domain model is initially trained on suboptimal demonstrations.

\section{Conclusion and Future Work}
\vspace{-0.15em}

\textbf{Conclusion.} We present SILVR, a self-improvement loop for solving novel robotic tasks via visual planning.  Initialized from an in-domain video model pretrained on a small general set of demonstrations, SILVR iteratively updates the visual planner for a novel task of interest through self-collected online experience.  Compared to equivalent behavior cloning setups, or utilizing online experience through reinforcement learning finetuning, we find that SILVR achieves superior self-improvement capabilities and demonstrates sample efficiency.  %
We successfully apply SILVR as a self-improving visual planner not only for synthetic environments, but also on robot arms deployed in the real world.

\textbf{Limitations and Future Work.} SILVR implicitly assumes that the initial in-domain model, optionally adapted with an internet-pretrained video model, achieves a reasonable success rate to collect online experience and achieve self-improvement. This assumption may not hold when the novel task is too challenging. Additionally, the choice of internet-pretrained video model can pose a trade-off on between video quality and computation cost. Whereas in this work we choose AnimateDiff~\citep{guo2023animatediff} as a large-scale pretrained video model with both reasonable generation quality as well as computational efficiency, more recent video generative models with enhanced visual quality can be further explored for improved visual planning performance on downstream robotic tasks.

Whereas SILVR achieves the best self-improvement performance with some initial success rate on the novel task, the cold start problem may pose challenges.  Just as exploration can help address the cold start problem for standard reinforcement learning, we believe that investigating how to improve exploration for the visual planning framework in a principled manner is a promising future direction.

\textbf{Acknowledgments.} This work is partially supported by Samsung and the U.S. National Science Foundation under Cooperative Agreement No. 2433429. Our research was conducted using computational resources at the Center for Computation and Visualization at Brown University. We would like to thank Professors George Konidaris and Stefanie Tellex for their generous support for our real-robot experiments. We would also like to thank Shijie Wang and Zitian Tang for their help with the experiments. Calvin would like to thank Dennis Lee and Ally Koo, as well as Kayan Shih and her family, for all their kindness and hospitality during the paper writing process.

\section{Reproducibility Statement}
 To support reproducibility, all codebases utilized in this paper were modified versions of publicly available repositories.  Furthermore, we provide extensive details of hyperparameters and settings both in the main body of the text as well as in the Appendix.

\clearpage
\appendix

\setcounter{figure}{0}
\renewcommand{\thefigure}{A\arabic{figure}}
\renewcommand{\thetable}{A\arabic{table}}

\section{Tasks and Text Prompts}
Below we list the tasks and associated text prompts used for evaluating SILVR.  Tasks with demonstrations seen during training of the in-domain model are denoted with an asterisk.

\begin{table}[h]
\centering
\small

\setlength{\tabcolsep}{4.8pt}
\scalebox{0.8}{

\begin{tabular}{@{}lll@{}}
\toprule
Task             & In-Domain Model Prompts                         & Internet-Domain Model Prompts                                                         \\ \midrule
Coffee Button$^*$     & coffee button                         &  a robot arm pressing the coffee machine button                               \\
Door Open$^*$     & door open                         & a robot arm reaching a door handle and pulling it to open the door                                                                    \\
Drawer Open$^*$     & drawer open                         & a robot arm opening a drawer by pulling its white handle backward                                                                      \\
Peg Unplug Side$^*$     & peg unplug side                         & a robot arm unplugging a peg by pulling it from the right to the left                                                                     \\
Plate Slide Side$^*$       & plate slide side                           & a robot arm sliding a plate from the left into the net on the right side                                                             \\ 
Push$^*$      & push                          & a robot arm pushing an object forward to a green sphere                                       \\
Reach$^*$     & reach & a robot arm reaching a red sphere \\
Sweep$^*$    & sweep & a robot arm moving an object to the left side of the table \\
Door Close       & door close                           & a robot arm pushing a door to close it                                                                 \\
Window Close     & window close                         & a robot arm closing a window by pulling its handle from the left to the right                                                                      \\
Window Open      & window open                          & a robot arm opening a window by pushing its handle from the right to the left                                                                      \\
Drawer Close     & drawer close                         & a robot arm closing a drawer by pushing its white handle forward                                                                      \\
Faucet Close     & faucet close                         & a robot arm pulling a faucet counterclockwise                                                                      \\
Faucet Open      & faucet open                          & a robot arm pushing a faucet clockwise                                                                      \\
Handle Press     & handle press                         & a robot arm pressing down a handle                                                                      \\
Handle Press Side     & handle press side                & a robot arm pressing down a handle on the side                                                                      \\
Dial Turn     & dial turn                          & a robot arm turning a dial counterclockwise                                                                      \\
Plate Slide     & plate slide                          & a robot arm sliding a plate forward into the net                                                                      \\
Reach Wall     & reach wall                          & a robot arm reaching toward a red sphere over a wall                                                                      \\
Button Press Wall    & button press wall                         & a robot arm reaching over a wall to press a button                                                                      \\ \midrule
Push Red Cup$^*$     & red                         & a robot arm pushing the red cup                                                                      \\
Push Blue Cup$^*$     & blue                         & a robot arm pushing the blue cup                                                                      \\
Push Green Cup$^*$     & green                         & a robot arm pushing the green cup                                                                      \\
Push Pink Cup$^*$     & pink                         & a robot arm pushing the pink cup                                                                      \\
Push Orange Cup     & orange                         & a robot arm pushing the orange cup                                                                      \\
Push Purple Cup     & purple                         & a robot arm pushing the purple cup       \\
\midrule
Open Red Drawer$^*$     & red                         & a robot arm opening the red drawer                                                                      \\
Open Green Drawer$^*$     & green                         & a robot arm opening the green drawer                                                                      \\
Open Blue Drawer$^*$     & blue                         & a robot arm opening the blue drawer                                                                     \\
Open Yellow Drawer     & yellow                         & a robot arm opening the yellow drawer
\\\bottomrule
\end{tabular}
}
\vspace{1em}
\caption{\textbf{Task-Prompt Pairs.} We include a comprehensive list of tasks and their text prompts for in-domain training and evaluation. ``$*$'' denotes tasks seen during initial training of the in-domain model.  We also provide the prompts used to interface with the internet-pretrained text-to-video model during adaptation with IPA.}
\label{table:text_prompts}
\end{table}

\section{Implementation Details}
\label{sec:implement_appendix}
We provide detailed architecture configurations of the models used in SILVR, and their relevant hyperparameter settings below.

\textbf{MLP Inverse Dynamics Model:} Following prior work~\citep{luo2025solving}, we design one choice of inverse dynamics model as a small MLP network built on top of a pretrained pixel-based representation network.  The MLP-IDM takes as input the embeddings of two video frames, which are extracted using VC-1~\citep{majumdar2023vc1}, and outputs a prediction of the action that enables the transition between the provided frames.

For the Panda arm experiments, the MLP-IDM is tasked with predicting the end effector position of the last frame provided.  This is then executed in the physical environment through inverse kinematics.  Furthermore, the two video frames have a frameskip of 16; the frequency at which the camera is queried for trajectories is so high such that two temporally consecutive frames is not more substantially meaningful than just observing the last frame.  For MetaWorld experiments, the two video frames are consecutive, and thus have a frameskip of 1.

The total parameter count of the MLP-IDM used in real-world experimentation is 85.81M. Of these, 85.80M parameters are inherited from VC-1 whereas our MLP-IDM design contributes only an additional 10759 parameters due to the additional MLP on top.

In our real-world experiments, we reuse the same MLP-IDM for all tasks within the same environments, and do not perform any finetuning during the SILVR iterations with subsequently self-collected data.  In such a way, the MLP-IDM is trained on a set of seen tasks, but applied to a novel task without further modification.  The subsequent success on such novel tasks therefore highlights not only the robustness of the MLP-IDM learned, but also the visual quality of the synthesized visual plans. The detailed hyperparameters of MLP-IDM training are provided in Table~\ref{table:inv_dyn_hparams}.

\begin{table}[ht]
\centering
\small

\begin{tabular}{@{}ll@{}}
\toprule
Hyperparameter    & Value \\ \midrule
Input Dimension  & 1536      \\
Output Dimension (Panda) & 7      \\
Training Epochs  & 20      \\
Learning Rate    & 1e-5      \\
Optimizer        & AdamW     \\ \bottomrule
\end{tabular}

\vspace{1em}
\caption{\textbf{Hyperparameters of MLP Inverse Dynamics Model Training.}  We list the relevant hyperparameters of training the MLP inverse dynamics model.}
\label{table:inv_dyn_hparams}
\end{table}

\textbf{Diffusion Inverse Dynamics Model (DIDM):} Whereas the MLP-IDM does continue to facilitate self-improvement for MetaWorld experiments through SILVR, we find that the most performant implementation for our simulated experiments was a Diffusion Inverse Dynamics Model (DIDM).  The DIDM is built off the UNet implementation of a Diffusion Policy~\citep{chi2023diffusionpolicy}; it is modified to take in not only the current frame but also a frame 9 timesteps into the future and outputs an action chunks of size 8.  We further implement the DIDM to operate directly in StableDiffusion latent space, where each frame is of dimension (64,64,4), rather than RGB space of (512,512,3) for further speed efficiency.  The DIDM is initially trained on the 8 seen-task set (with 25 initial demonstrations per task) for 200 epochs, with a learning rate of 1e-4, and a batch size of 128.  At each SILVR iteration on MetaWorld, the DIDM is further finetuned on the 30 collected demonstrations for 30 epochs, reusing a lr of 1e-4, and with a batch size of 30.

\begin{table}[ht]
\centering
\small

\begin{tabular}{@{}ll@{}}
\toprule
Hyperparameter    & Value \\ \midrule
Input Dimension  & 32768      \\
Output Dimension (MetaWorld) & 4      \\
Training Epochs  & 200      \\
Learning Rate    & 1e-4      \\
Optimizer        & AdamW     \\ \bottomrule
\end{tabular}

\vspace{1em}
\caption{\textbf{Hyperparameters of Diffusion Inverse Dynamics Model Training.}  We list the relevant hyperparameters of training the diffusion inverse dynamics model.}
\label{table:diff_dyn_hparams}
\end{table}

\textbf{In-Domain Model:} We reuse the implementation of a small-scale diffusion model that conditions on both natural language and an initial pixel frame from~\citep{ko2024avdc}.  To improve text-conditioned capabilities of the model, we add an additional Cross-Attention layer to every level of the U-Net, which attends to the CLIP-encoded text prompt. Specifically, we instantiate UNet with 3 ResNet blocks for MetaWorld settings and 2 ResNet blocks for Panda arm tasks. We report the detailed list of model parameters in Table~\ref{table:avdc_params}. In total, the in-domain model consists of 179.91M parameters for MetaWorld and 156.58M parameters for Real-World experiments. We perform initial in-domain training for 70K training steps on MetaWorld and 88K steps on Panda, with a batch size of 8 and a learning rate of 2e-5. In each SILVR iteration, we finetune the in-domain video model for 10K steps with with a batch size of 4 and a learning rate of 1e-6 on MetaWorld. On Panda Arm, we finetune for 8,000 steps with a batch size of 8 and a learning rate of 2e-5 on Cup Pushing and for 10,000 steps with a batch size of 8 and a learning rate of 1e-5 on Drawer Opening. All experiments are performed on a single NVIDIA A6000 or RTX3090 GPU.

\begin{table}[ht]
\centering
\setlength{\tabcolsep}{4.8pt}
\scalebox{0.9}{
\begin{tabular}{@{}ll@{}}
\toprule
Component   & \# Parameters (Millions) \\
\midrule
U-Net (MetaWorld) & 116.71 \\
U-Net (Panda Arm) & 93.38 \\
Text Encoder (\texttt{openai/clip-vit-base-patch32}) & 63.2 \\
\bottomrule
\end{tabular}
}

\vspace{1em}
\caption{\textbf{In-Domain Model Components.} SILVR relies on a small in-domain text-to-video model, which we base our implementation off of prior work~\citep{ko2024avdc}.  We list the size of the components of the model architecture used.}
\label{table:avdc_params}
\end{table}

\textbf{Internet-Domain Model:} Following Adapt2Act~\citep{luo2025solving}, we employ AnimateDiff~\citep{guo2023animatediff} as the frozen internet-pretrained video model for inverse probabilistic adaptation. Additionally, we use SparseCtrl~\citep{Guo2024ECCV_SparseCtrl_Adding_Sparse} to enable image-conditioned video generation. Model components and their parameter counts are listed in Table~\ref{table:animatediff_params}. In total, AnimateDiff consists of 2.005B parameters.

\begin{table}[ht]
\centering
\setlength{\tabcolsep}{4.8pt}
\scalebox{0.9}{
\begin{tabular}{@{}ll@{}}
\toprule
Component   & \# Parameters (Millions) \\
\midrule
VAE (Encoder) & 34.16 \\
VAE (Decoder) & 49.49 \\
U-Net & 1302.16 \\
Text Encoder & 123.06 \\
ControlNet & 496.73 \\
\bottomrule
\end{tabular}
}

\vspace{1em}
\caption{\textbf{AnimateDiff Components.} IPA relies on a internet-scale text-to-video model; in this work we use AnimateDiff.  We thus list the size of components of the AnimateDiff checkpoint used.  The checkpoint is used purely for inference, and is not modified or updated in any way. Note that the VAE Decoder is not utilized in our framework.}
\label{table:animatediff_params}
\end{table}

\textbf{Visual Planning Hyperparameters:} In visual planning, we predict 8 future frames conditioned on the current observation and task prompt. We follow~\citep{luo2025solving} to perform DDIM~\citep{song2021ddim} sampling for 25 steps to synthesize visual plans, in which the text-conditioning guidance scale is set to 2.5 for MetaWorld experiments and 7.0 for Panda Arm Pushing. We use 0.5 as the prior strength for inverse probabilistic adaptation.

\textbf{Choices of Control Loop:} Visual planning provides the user control over the quality of execution against the speed. In our experiments, each visual plan consists of 9 frames, including one current observation and eight future frames, and can be translated into 8 actions. By performing open-loop control, we execute all 8 actions from a single visual plan sequentially in the environment without any re-planning. While synthesizing a visual plan can often involve multiple sampling steps and thus be time-consuming, open-loop control greatly improves the interaction efficiency. However, since open-loop control does not adjust the control actions based on the feedback from the environment, the subsequent actions from the plan might become suboptimal to the latest states and cause error accumulation. To mitigate this issue, closed-loop control adjusts the action for every interaction step. Specifically, we execute only the first action from the plan, and perform re-planning based on the new observation received from the environment. Although this control style allows us to interact most reliably, it incurs a large computational overhead due to frequent re-planning. To balance the execution quality and efficiency, we can flexibly choose a control loop between the two extremes of open-loop and closed-loop. For example, we execute half of the plan (e.g. 4 actions) before re-planning, which we reference as semi-open-loop control.

To achieve the best execution speed, we employ open-loop control in Panda Arm Pushing and Drawer Opening tasks, in which we discover that visual plans can be performed decently, with negligible deviation in the real execution. For all MetaWorld experiments, we utilize semi-open-loop control to balance performance and efficiency.

\textbf{Behavior Cloning Improvement Loop}: A Diffusion Policy is initially trained on the same data as used in SILVR for 150 epochs with learning rate 1e-4 and batch size 64. For each iteration, we deploy the diffusion policy to interact with the environment and collect 30 task demonstrations. The policy will then be fine-tuned on the successful data filtered by ground-truth task success signals for 50 epochs with batch size 30 and learning rate 1e-4. 

\textbf{SILVR-Distilled Diffusion Policy}: After applying SILVR, we optionally distill the visual planning components into a diffusion policy.  This has the benefit of lightweight, fast inference for final deployment, while still leveraging the self-improving benefits of visual models during the training process, thus balancing both worlds.  The architecture of the DP is consistent with that used in the DSRL and BCIL experiments.  During distillation, the visual planner teacher model first collects 120 demonstrations from the environment.  Then, a DP is trained for 300 epochs with a batch size of 64 and a learning rate of 1e-4 on only these 120 demonstrations collected by the teacher.

In essence, SILVR can be thought of as composed of two systems.  The slower system is the video planner approach, where the advantage appears in greater autonomous improvement capabilities.  The faster system is the distilled diffusion policy, which although has weaker self-improvement, can capture the current performance of the slower system and be deployed with fast inference.

\textbf{DSRL Implementation}: We utilize the open-source implementation of DSRL~\citep{wagenmaker2025steering} for our experiments.  We maintain the vast majority of parameters from the default setting, but make sure to collect 30 demonstrations for each iteration as in SILVR. When optimizing over the collected experience, we utilize a comparable amount of gradient steps with the amount found in the default settings of DSRL; we use a batch size of 256 with 60000 gradient steps per update.  We find that given the same amount of experience and a sufficiently large update budget, DSRL is unable to match the iterative performance improvements of SILVR.  This highlights the sample inefficiency of reinforcement learning, which may need extra experience to successfully bootstrap a value function for the novel task of interest.

We performed a light ablation, comparing a roughly equivalent amount of gradient updates with SILVR (150 gradient steps per update) against the amount of updates used across multiple standard robotic tasks provided by the DSRL publicly available codebase, and found no significant difference in improvement trend.  This suggests that the bottleneck for self-improvement through DSRL is not update computation, but experience collection; on the other hand, SILVR demonstrates significantly better sample efficiency for iterative self-improvement.

\begin{table}[h]
\centering
\begin{tabular}{@{}llllll@{}}
\toprule
Iteration & 0           & 1           & 2           & 3           & 4           \\ \midrule
DSRL (150 Updates)   & 10.1 (±0.1)  & 8.9 (±0.6)   & 8.6 (±1.0)   & 9.4 (±0.1)   & 8.3 (±0.2)   \\
DSRL (60000 Updates) & 9.4 (±1.7)   & 8.3 (±1.6)   & 7.4 (±0.9)   & 7.5 (±3.4)   & 7.7 (±3.4)   \\ \bottomrule
\end{tabular}
\caption[]{\textbf{DSRL Performance Ablation over Different Update Rates.} We report mean success rate and standard deviation across three iterations for DSRL with different numbers of updates per iteration. We show that the bottleneck for DSRL is not gradient updates, but collected experience.}
\label{table:dsrl_update_comparison}
\end{table}

\vspace{-1em}
\section{MetaWorld Task Performance Decomposition}

\begin{table}[h!]
\centering
\setlength{\tabcolsep}{4.8pt}
\scalebox{0.58}{
\begin{tabular}{@{}lllllllllll@{}}
\toprule
Iteration         & 0                            & 1                             & 2                             & 3                             & 4                             & 5                             & 6                             & 7                             & 8                             & 9                             \\ \midrule
Button Press Wall & 0.0 ($\pm$0.0)  & 0.0 ($\pm$0.0)   & 0.0 ($\pm$0.0)   & 0.0 ($\pm$0.0)   & 0.0 ($\pm$0.0)   & 0.0 ($\pm$0.0)   & 0.0 ($\pm$0.0)   & 0.0 ($\pm$0.0)   & 0.0 ($\pm$0.0)   & 0.0 ($\pm$0.0)   \\
Dial Turn         & 4.4 ($\pm$1.9)  & 25.6 ($\pm$22.7) & 17.8 ($\pm$15.8) & 23.3 ($\pm$20.8) & 24.4 ($\pm$19.0) & 25.6 ($\pm$22.2) & 28.9 ($\pm$25.9) & 36.7 ($\pm$31.8) & 36.7 ($\pm$32.8) & 34.4 ($\pm$30.2) \\
Door Close        & 35.6 ($\pm$6.9) & 57.8 ($\pm$26.9) & 58.9 ($\pm$25.2) & 74.4 ($\pm$15.8) & 72.2 ($\pm$18.4) & 80.0 ($\pm$13.3) & 81.1 ($\pm$11.7) & 83.3 ($\pm$8.8)  & 78.9 ($\pm$12.6) & 85.6 ($\pm$5.1)  \\
Drawer Close      & 1.1 ($\pm$1.9)  & 11.1 ($\pm$7.7)  & 20.0 ($\pm$6.7)  & 21.1 ($\pm$6.9)  & 20.0 ($\pm$12.0) & 31.1 ($\pm$10.7) & 27.8 ($\pm$10.7) & 27.8 ($\pm$13.5) & 28.9 ($\pm$19.0) & 31.1 ($\pm$13.5) \\
Faucet Close      & 17.8 ($\pm$6.9) & 27.8 ($\pm$25.2) & 41.1 ($\pm$20.1) & 50.0 ($\pm$24.0) & 66.7 ($\pm$26.5) & 67.8 ($\pm$13.9) & 73.3 ($\pm$16.7) & 80.0 ($\pm$13.3) & 73.3 ($\pm$17.3) & 77.8 ($\pm$3.8)  \\
Faucet Open       & 6.7 ($\pm$6.7)  & 24.4 ($\pm$18.4) & 37.8 ($\pm$6.9)  & 57.8 ($\pm$6.9)  & 60.0 ($\pm$8.8)  & 82.2 ($\pm$1.9)  & 78.9 ($\pm$10.2) & 80.0 ($\pm$10.0) & 71.1 ($\pm$11.7) & 77.8 ($\pm$10.2) \\
Handle Press      & 35.6 ($\pm$7.7) & 41.1 ($\pm$10.2) & 40.0 ($\pm$3.3)  & 56.7 ($\pm$8.8)  & 58.9 ($\pm$13.5) & 56.7 ($\pm$3.3)  & 63.3 ($\pm$5.8)  & 70.0 ($\pm$5.8)  & 72.2 ($\pm$9.6)  & 67.8 ($\pm$5.1)  \\
Handle PressSide  & 18.9 ($\pm$5.1) & 31.1 ($\pm$12.6) & 48.9 ($\pm$19.0) & 72.2 ($\pm$13.5) & 60.0 ($\pm$8.8)  & 70.0 ($\pm$14.5) & 75.6 ($\pm$8.4)  & 66.7 ($\pm$6.7)  & 68.9 ($\pm$18.4) & 65.6 ($\pm$11.7) \\
Plate Slide       & 17.8 ($\pm$5.1) & 26.7 ($\pm$16.7) & 33.3 ($\pm$17.3) & 47.8 ($\pm$15.8) & 52.2 ($\pm$11.7) & 70.0 ($\pm$5.8)  & 60.0 ($\pm$5.8)  & 62.2 ($\pm$18.4) & 71.1 ($\pm$15.0) & 66.7 ($\pm$12.0) \\
Reach Wall        & 10.0 ($\pm$5.8) & 32.2 ($\pm$1.9)  & 22.2 ($\pm$9.6)  & 30.0 ($\pm$6.7)  & 27.8 ($\pm$5.1)  & 35.6 ($\pm$13.9) & 26.7 ($\pm$3.3)  & 24.4 ($\pm$8.4)  & 26.7 ($\pm$12.0) & 30.0 ($\pm$3.3)  \\
Window Close      & 3.3 ($\pm$3.3)  & 10.0 ($\pm$0.0)  & 25.6 ($\pm$36.0) & 14.4 ($\pm$13.5) & 16.7 ($\pm$15.3) & 20.0 ($\pm$15.3) & 21.1 ($\pm$13.5) & 20.0 ($\pm$12.0) & 34.4 ($\pm$13.5) & 25.6 ($\pm$8.4)  \\
Window Open       & 25.6 ($\pm$5.1) & 44.4 ($\pm$18.4) & 56.7 ($\pm$11.5) & 74.4 ($\pm$1.9)  & 71.1 ($\pm$13.5) & 74.4 ($\pm$7.7)  & 81.1 ($\pm$11.7) & 73.3 ($\pm$8.8)  & 72.2 ($\pm$15.0) & 74.4 ($\pm$3.8)  \\ \midrule
AVG               & 14.7 ($\pm$0.6) & 27.7 ($\pm$1.9)  & 33.5 ($\pm$2.2)  & 43.5 ($\pm$2.6)  & 44.2 ($\pm$4.5)  & 51.1 ($\pm$4.0)  & 51.5 ($\pm$3.0)  & 52.0 ($\pm$5.0)  & 52.9 ($\pm$5.2)  & 53.1 ($\pm$3.8)  \\ \bottomrule
\end{tabular}
}
\vspace{1em}
\caption[]{\textbf{MetaWorld Task Performance.} We provide a detailed list of task performance for the leftmost plot in Figure~\ref{fig:silvr_iter10}. We report the mean success rate and standard deviation aggregated over $3$ seeds each, across 10 iterations.  We reiterate that none of these 12 tasks had been seen a priori during initial in-domain video model training.}
\label{table:silvr_expert_task_perf_decomposition}
\end{table}

\section{Generalization Capabilities between Visual Planning and Action-Predictive BC}

Our results suggest that visual planning has some default generalization and self-improvement benefits for decision-making compared to using direct action-prediction BC policies.  We hypothesize that modeling consistent visual motions can extract more training signal from the provided data than modeling action sequences directly, as there is more supervisory training signal from pixels than low-dimensional actions.  Indeed, as shown in the MetaWorld portion of Figure~\ref{fig:franka_sail_visual_plans_comparison_all_tasks}, for initial iteration 0 on a novel task concerning an unseen object, the visual planner still manages to generate coherent motions for the robot arm even if it blurs out the specific novel object interaction. These coherent motions, even if the specific object interaction is not modeled correctly (or even blurred) initially, can still be accurately translated by the IDM into meaningful robot actions; thus visual planning may have additional generalization benefits. On the other hand, the basic BC policy does not model visual details but predicts a sequence of actions entirely from the conditioning frame; having overfit to its training set, when faced with a novel object in the scene it may predict highly suboptimal actions, thus leading to poorer generalization.

\section{Full MetaWorld Suboptimal Results}

\begin{figure}[ht]
    \centering
    \includegraphics[width=0.9\linewidth]{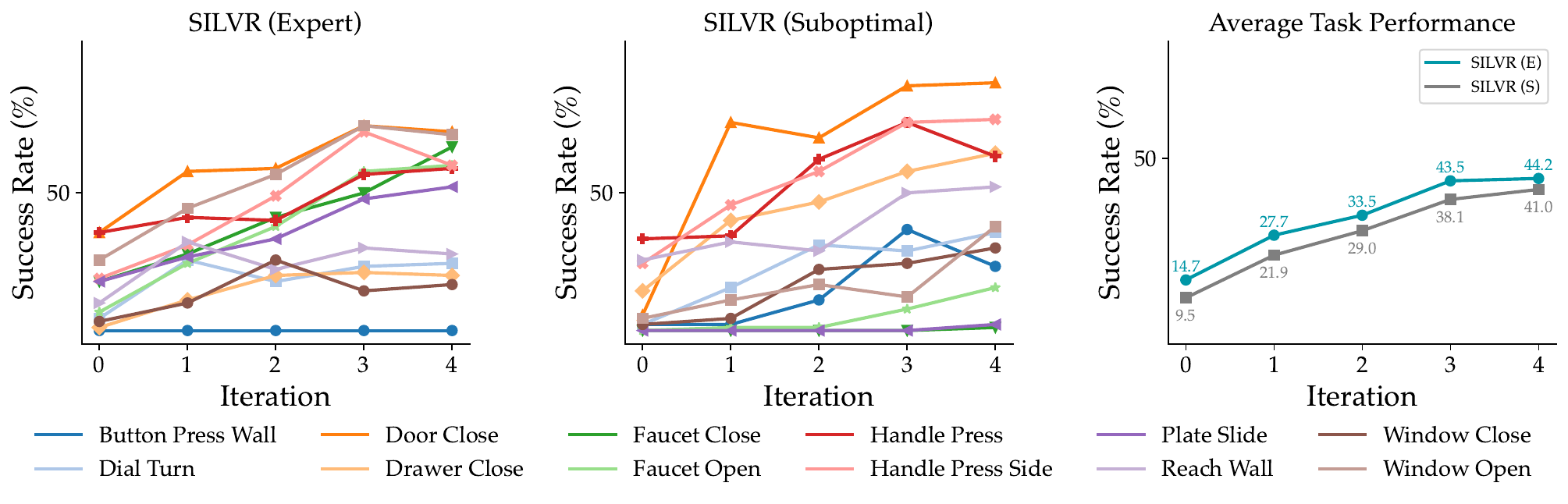}
    \caption{\textbf{SILVR Curves with suboptimal in-domain data.} For each task, we plot the mean success rate across 5 iterations, aggregated over $3$ seeds.}
    \label{fig:silvr_suboptimal_task_perf_decomposition}
\end{figure}

As mentioned in Section~\ref{sec:suboptimal_sail}, we evaluate SILVR on 12 unseen MetaWorld tasks with suboptimal initial in-domain data. We provide a detailed task performance breakdown in  Figure~\ref{fig:silvr_suboptimal_task_perf_decomposition} and Table~\ref{table:silvr_suboptimal_detailed_task_perf}. We observe that most tasks, as well as average performance, exhibit an improving trend overall across iterations, demonstrating the robustness of SILVR to initial data quality.

\begin{table}[ht]
\centering
\setlength{\tabcolsep}{4.8pt}
\scalebox{0.9}{
\begin{tabular}{@{}llllll@{}}
\toprule
Iteration         & 0           & 1            & 2            & 3            & 4            \\ \midrule
Button Press Wall & 2.2 ($\pm$1.9)  & 2.2 ($\pm$3.8)   & 11.1 ($\pm$10.2) & 36.7 ($\pm$32.1) & 23.3 ($\pm$20.8) \\
Dial Turn         & 2.2 ($\pm$1.9)  & 15.6 ($\pm$12.6) & 31.1 ($\pm$13.5) & 28.9 ($\pm$23.4) & 35.6 ($\pm$11.7) \\
Door Close        & 5.6 ($\pm$3.8)  & 75.6 ($\pm$18.4) & 70.0 ($\pm$8.8)  & 88.9 ($\pm$7.7)  & 90.0 ($\pm$3.3)  \\
Drawer Close      & 14.4 ($\pm$6.9) & 40.0 ($\pm$21.9) & 46.7 ($\pm$20.3) & 57.8 ($\pm$29.1) & 64.4 ($\pm$28.3) \\
Faucet Close      & 0.0 ($\pm$0.0)  & 0.0 ($\pm$0.0)   & 0.0 ($\pm$0.0)   & 0.0 ($\pm$0.0)   & 1.1 ($\pm$1.9)   \\
Faucet Open       & 0.0 ($\pm$0.0)  & 1.1 ($\pm$1.9)   & 1.1 ($\pm$1.9)   & 7.8 ($\pm$8.4)   & 15.6 ($\pm$16.8) \\
Handle Press      & 33.3 ($\pm$5.8) & 34.4 ($\pm$6.9)  & 62.2 ($\pm$15.0) & 75.6 ($\pm$13.9) & 63.3 ($\pm$21.9) \\
Handle Press Side & 24.4 ($\pm$1.9) & 45.6 ($\pm$9.6)  & 57.8 ($\pm$10.2) & 75.6 ($\pm$20.4) & 76.7 ($\pm$5.8)  \\
Plate Slide       & 0.0 ($\pm$0.0)  & 0.0 ($\pm$0.0)   & 0.0 ($\pm$0.0)   & 0.0 ($\pm$0.0)   & 2.2 ($\pm$1.9)   \\
Reach Wall        & 25.6 ($\pm$5.1) & 32.2 ($\pm$6.9)  & 28.9 ($\pm$7.7)  & 50.0 ($\pm$6.7)  & 52.2 ($\pm$6.9)  \\
Window Close      & 2.2 ($\pm$1.9)  & 4.4 ($\pm$1.9)   & 22.2 ($\pm$12.6) & 24.4 ($\pm$21.7) & 30.0 ($\pm$23.3) \\
Window Open       & 4.4 ($\pm$1.9)  & 11.1 ($\pm$6.9)  & 16.7 ($\pm$12.0) & 12.2 ($\pm$6.9)  & 37.8 ($\pm$22.7) \\ \midrule
AVG               & 9.5 ($\pm$0.3)  & 21.9 ($\pm$0.9)  & 29.0 ($\pm$2.3)  & 38.1 ($\pm$4.9)  & 41.0 ($\pm$4.4)  \\ \bottomrule
\end{tabular}
}
\vspace{1em}
\caption[]{\textbf{SILVR Performance with Suboptimal Initial Data.} For each task, we report the mean success rate and standard deviation aggregated over $3$ seeds, across 5 iterations.}
\label{table:silvr_suboptimal_detailed_task_perf}
\end{table}

Additionally, we provide BCIL and DSRL performance with suboptimal in-domain data in Table~\ref{table:baseline_suboptimal_results}. Surprisingly, we observe that BCIL benefits more from suboptimal initialization. We hypothesize that the exploration brought by initial random actions is crucial for bootstrapping the Diffusion Policy performance in BCIL.

\begin{table}[h]
\centering
\begin{tabular}{@{}llllll@{}}
\toprule
Iteration & 0          & 1           & 2           & 3           & 4           \\ \midrule
DSRL      & 7.4 (±3.8) & 7.1 (±0.6)  & 8.6 (±1.3)  & 7.8 (±1.0)  & 8.1 (±0.8)  \\
BCIL      & 8.1 (±1.0) & 21.6 (±0.9) & 29.1 (±2.3) & 37.2 (±6.1) & 39.6 (±5.3) \\ \bottomrule
\end{tabular}
\caption[]{\textbf{Baseline Performance with Suboptimal Initial Data.} We report the mean success rate and standard deviation across 12 unseen tasks, aggregated over $3$ seeds each.}
\label{table:baseline_suboptimal_results}
\end{table}

\section{MetaWorld Component Ablations}

\begin{table}[h]
\centering
\begin{tabular}{@{}llllll@{}}
\toprule
Iteration & 0          & 1           & 2           & 3           & 4           \\ \midrule
Finetuning All (SILVR)      & 14.7 (±0.6) & 27.7 (±1.9)  & 33.5 (±2.2)  & 43.5 (±2.6)  & 44.2 (±4.5)  \\
No IDM Finetuning      & 14.1 (±1.6) & 22.2 (±3.1) & 24.6 (±2.5) & 24.8 (±1.7) & 26.8 (±1.9) \\
Only IDM Finetuning      & 15.0 (±1.8) & 24.4 (±3.2) & 27.7 (±3.5) & 26.9 (±4.1) & 29.8 (±3.4) \\ \bottomrule
\end{tabular}
\caption[]{\textbf{Component Update Ablations.} We report the mean success rate and standard deviation across 12 unseen tasks, aggregated over $3$ seeds each.}
\label{table:component_ablation_results}
\end{table}

We investigate the importance of finetuning different visual planning components with respect to online experience on final task performance in the MetaWorld suite.  We find that in contrast with the real-world experiments, keeping the IDM frozen for the MetaWorld suite struggles to adjust to novel objects and motions, resulting in modest performance gains even when the video model improves.  Similarly, we have found that when the IDM is finetuned but the video model is not, there are indeed some performance gains most likely from better translation of the base generalization ability of the video model; but performance also saturates quickly.  Thus, under situations with heavy novelty such as across objects and motions, updating both components with respect to online experience is critical. Meanwhile, in real-world experiments, we observed that the initially trained IDMs are inherently robust to novel tasks, in which many motions learned from seen tasks can be reused, and can be utilized directly to great effect without additional finetuning.

\newpage
\section{Additional Plan Visualizations}
We show additional visual plans for SILVR, across multiple environments and tasks, along with their execution results.
\label{sec:additional_qualitative_visuals}

\subsection{SILVR with Ground-Truth Filtering}
Visual plans and their executions for SILVR with Ground-Truth filtering are illustrated below.

\begin{figure}[ht]
    \centering
    \includegraphics[width=0.9\linewidth]{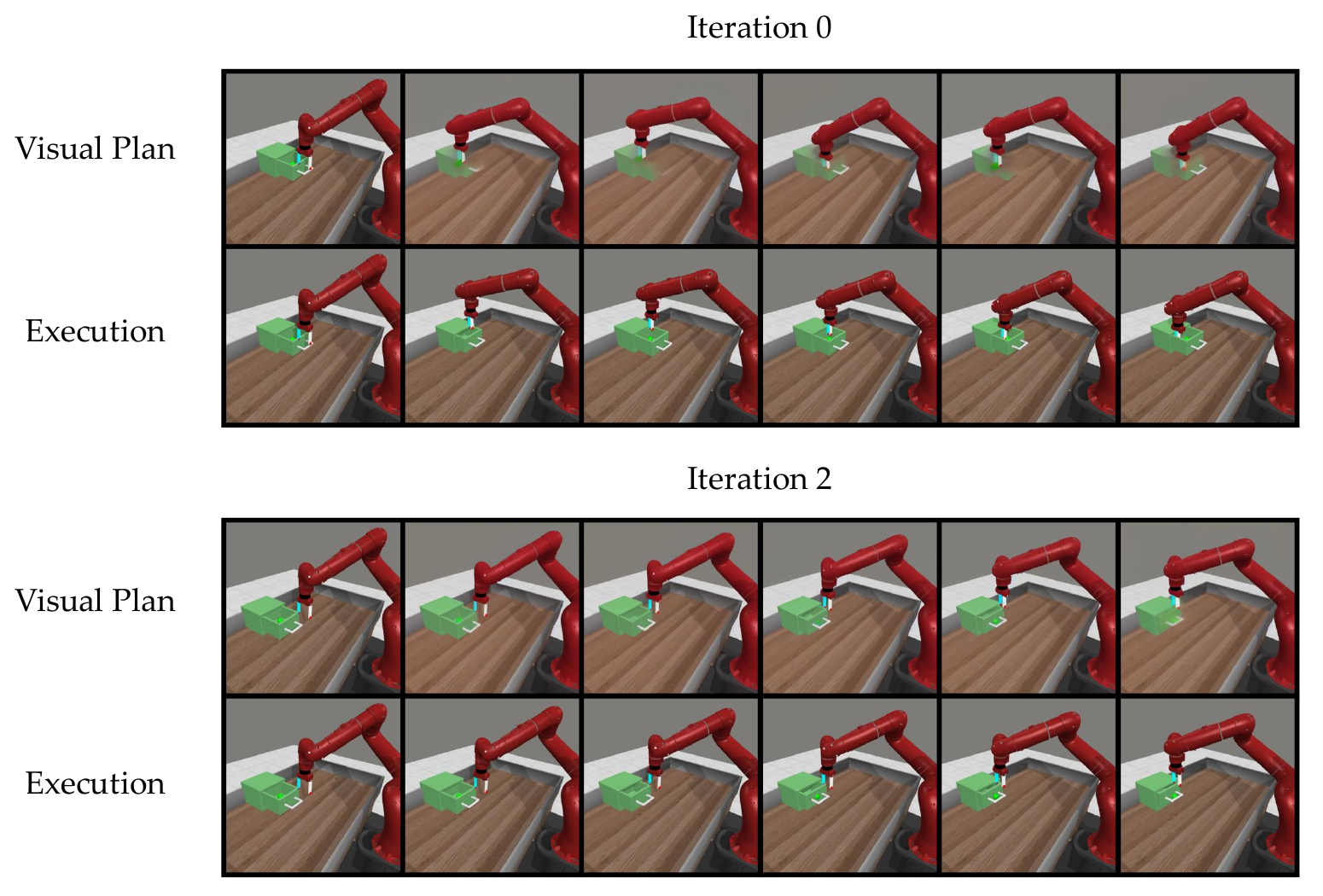}
    \caption{\textbf{SILVR on Drawer Close with Ground-Truth filtering.}}
    \label{fig:drawer_close_iter0_2}
\end{figure}

\begin{figure}[ht]
    \centering
    \includegraphics[width=0.9\linewidth]{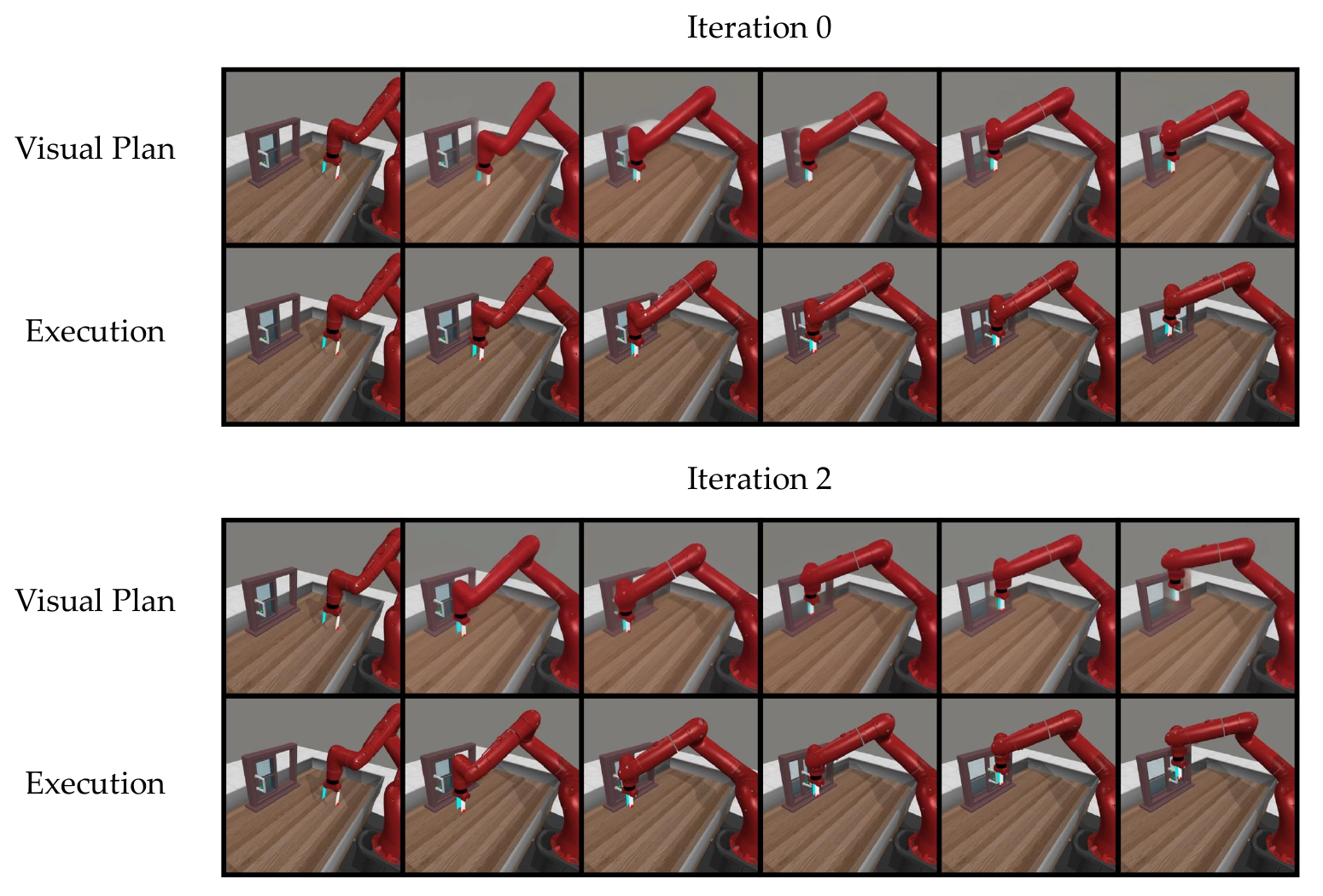}
    \caption{\textbf{SILVR on Window Close with Ground-Truth filtering.} }
    \label{fig:window_close_iter0_2}
\end{figure}

\begin{figure}[ht]
    \centering
    \includegraphics[width=\linewidth]{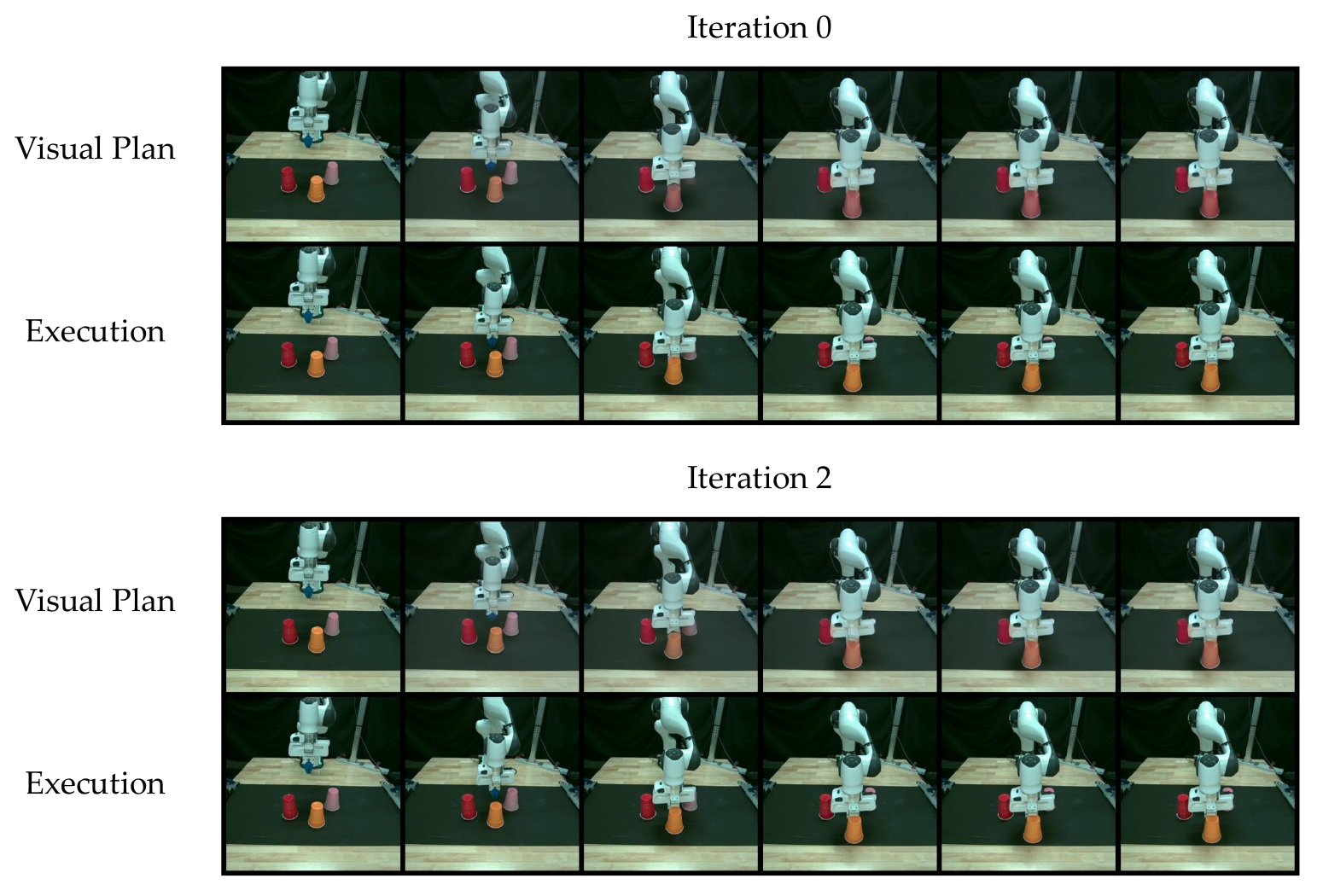}
    \caption{\textbf{SILVR on Orange Cup Pushing (Red/Pink/Orange) with Ground-Truth filtering.}}
    \label{fig:rpo_iter0_2}
\end{figure}

\begin{figure}[ht]
    \centering
    \includegraphics[width=\linewidth]{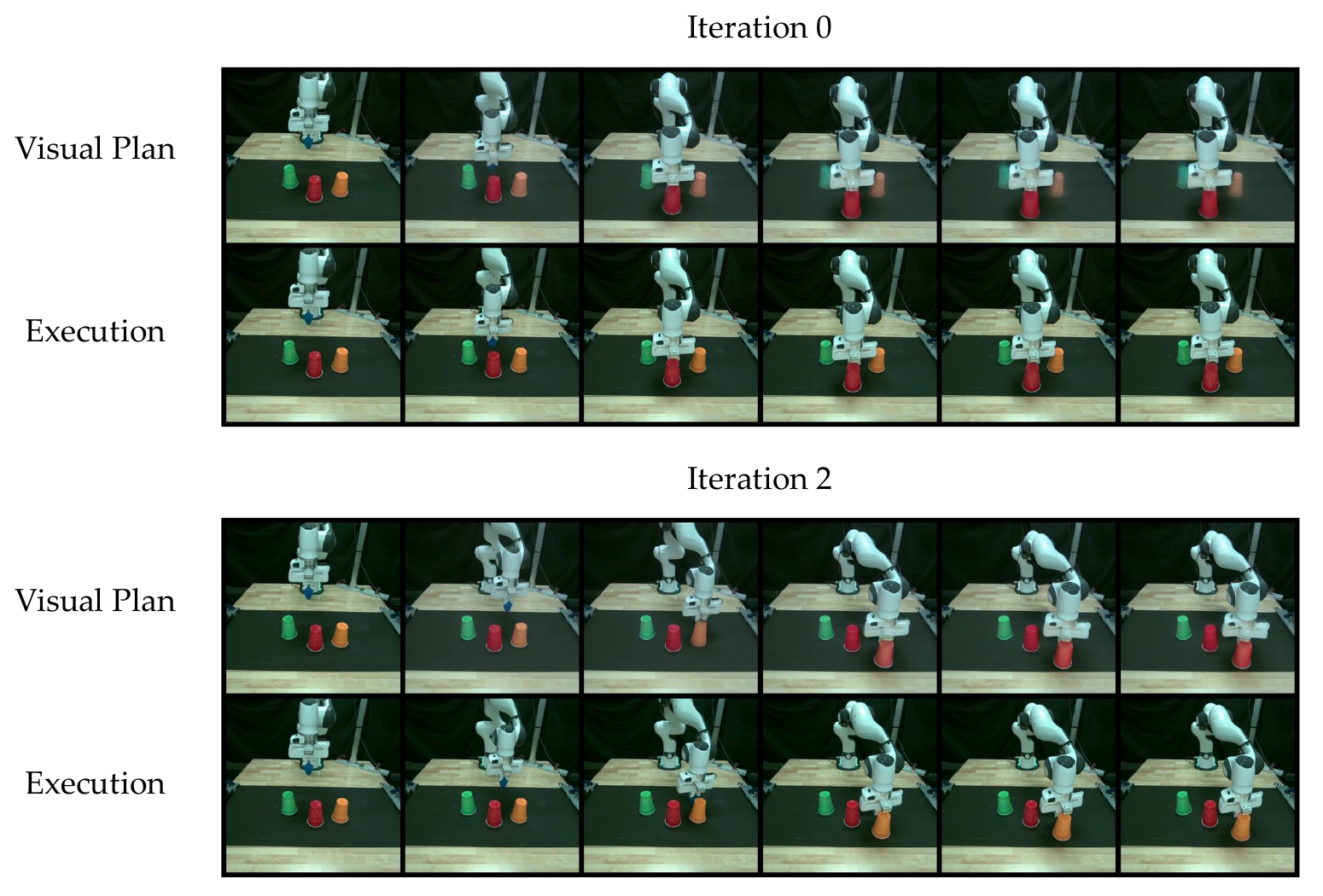}
    \caption{\textbf{SILVR on Orange Cup Pushing (Red/Green/Orange) with Ground-Truth filtering.}}
    \label{fig:rgo_iter0_2}
\end{figure}

\begin{figure}[ht]
    \centering
    \includegraphics[width=\linewidth]{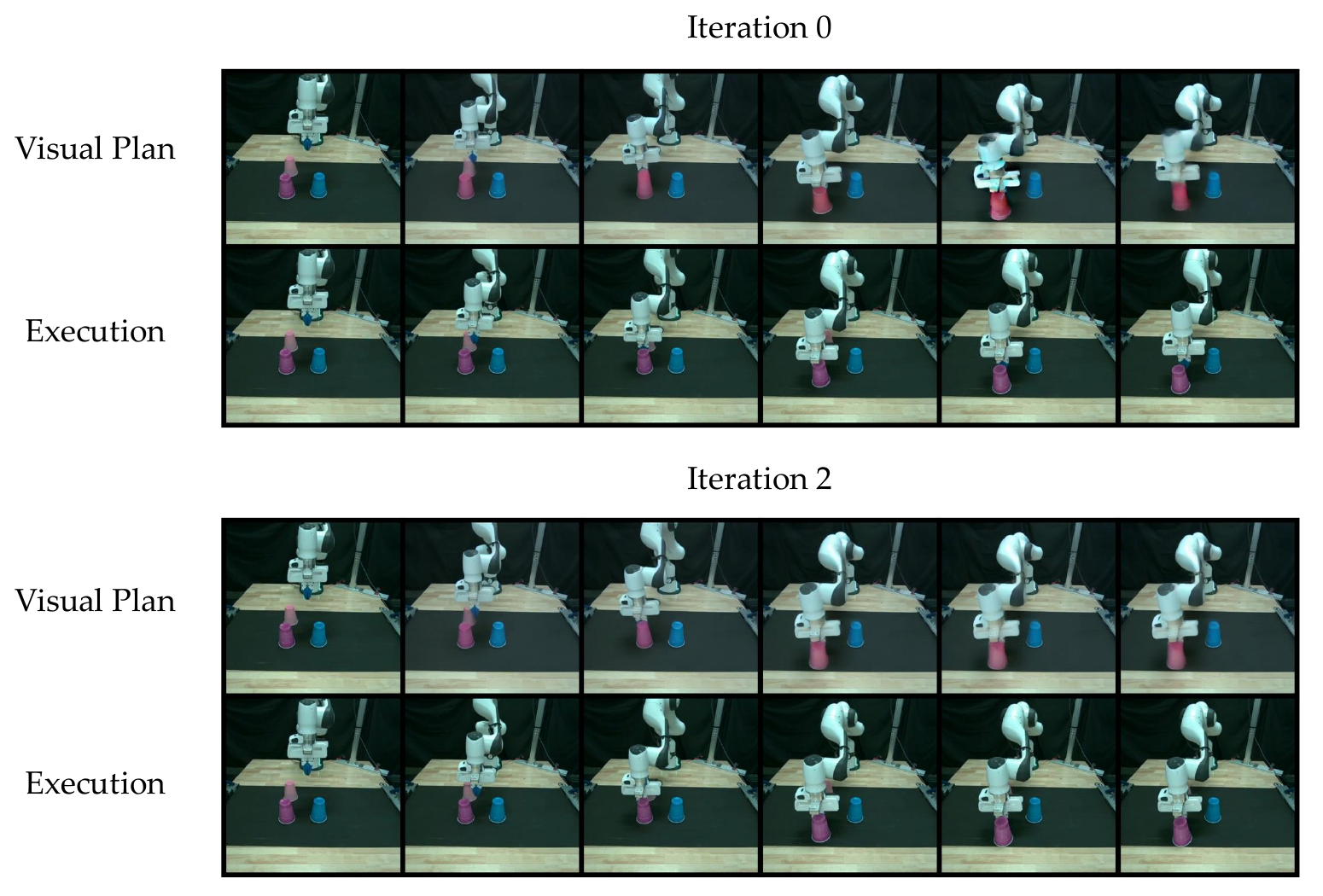}
    \caption{\textbf{SILVR on Purple Cup Pushing (Blue/Pink/Purple) with Ground-Truth filtering.}}
    \label{fig:bpp_iter0_2}
\end{figure}

\begin{figure}[ht]
    \centering
    \includegraphics[width=\linewidth]{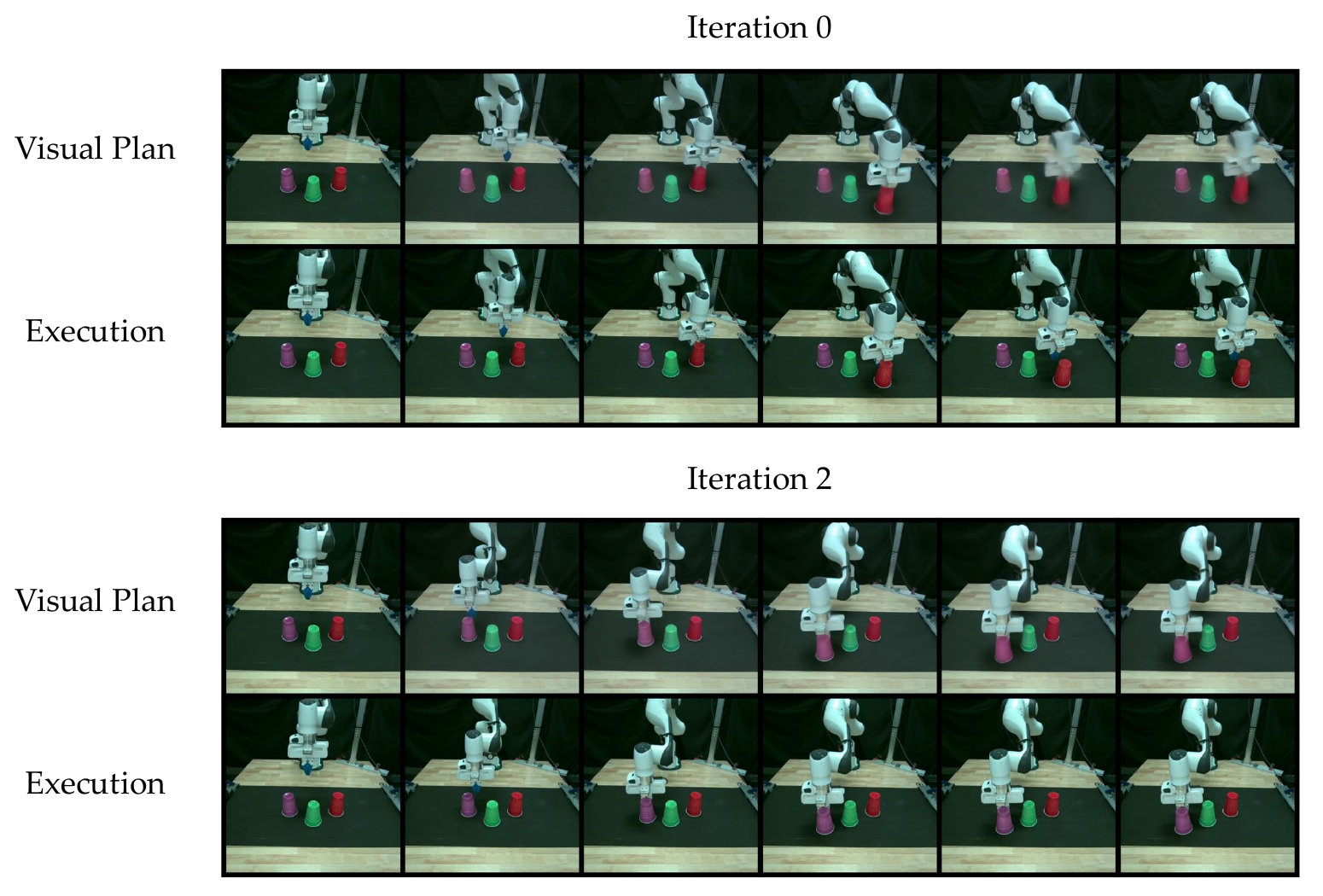}
    \caption{\textbf{SILVR on Purple Cup Pushing (Red/Green/Purple) with Ground-Truth filtering.}}
    \label{fig:rgp_iter0_2}
\end{figure}

\begin{figure}[ht]
    \centering
    \includegraphics[width=\linewidth]{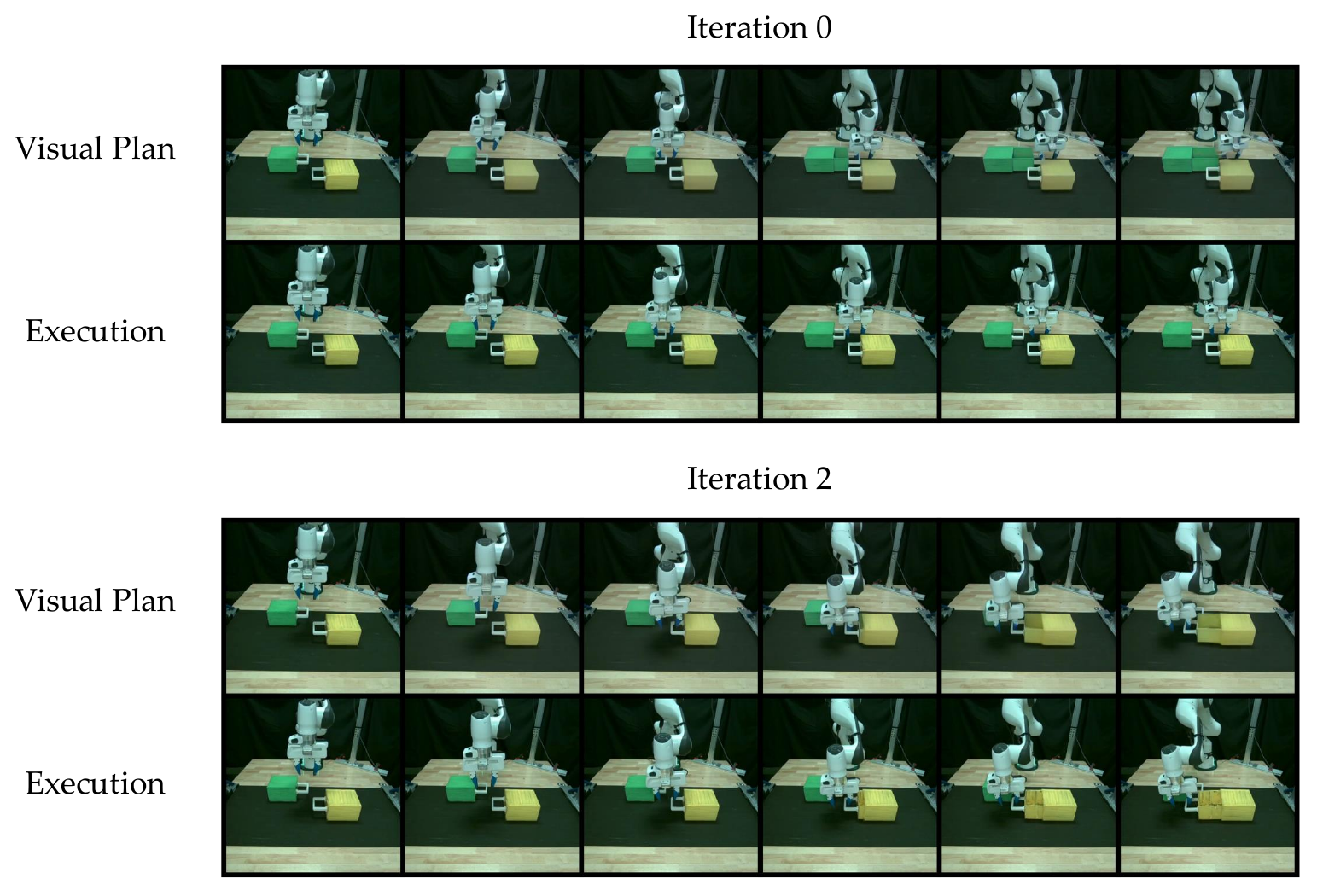}
    \caption{\textbf{SILVR on Yellow Drawer Opening (Yellow/Green) with Ground-Truth filtering.}}
    \label{fig:yg_iter0_2}
\end{figure}

\begin{figure}[ht]
    \centering
    \includegraphics[width=\linewidth]{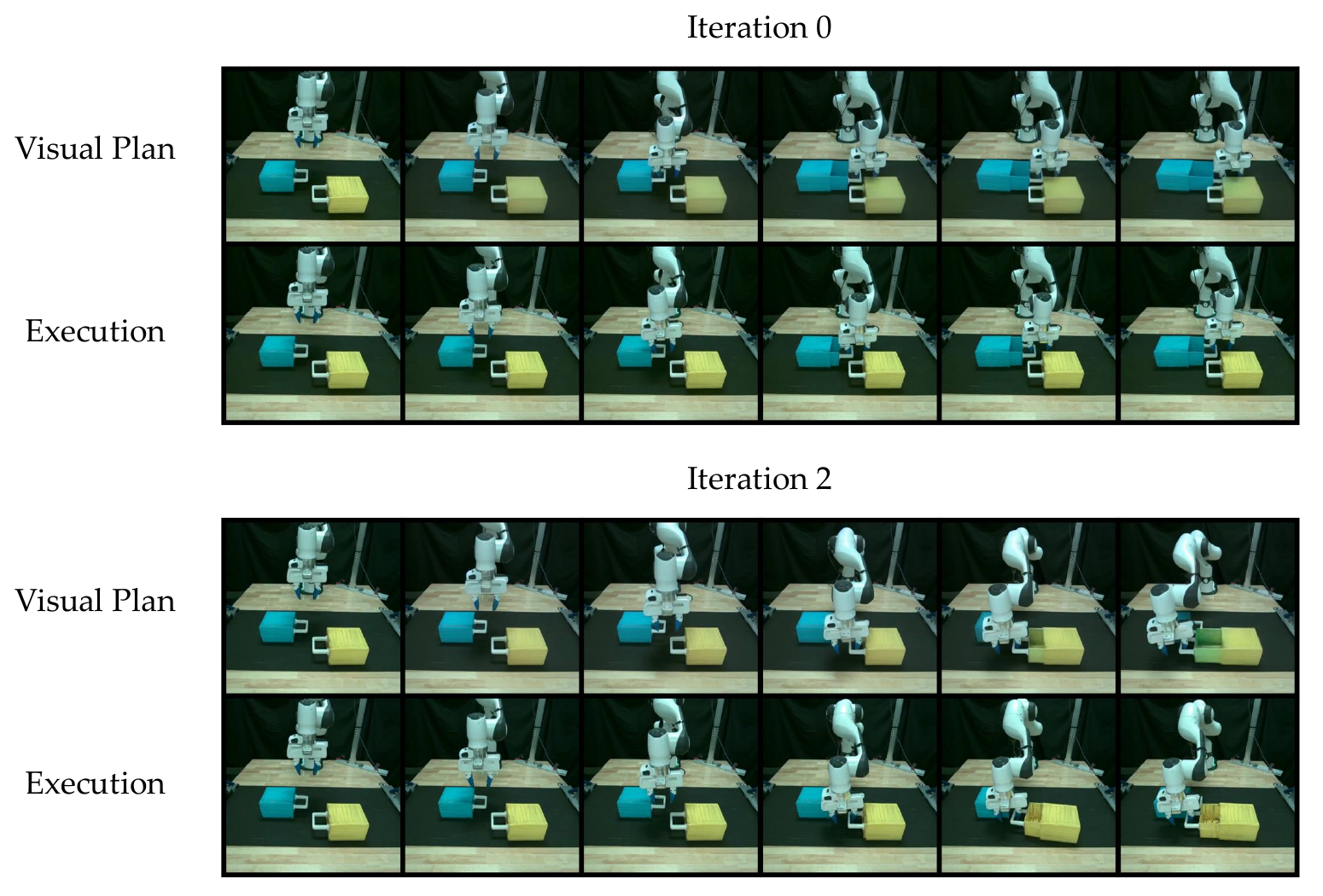}
    \caption{\textbf{SILVR on Yellow Drawer Opening (Yellow/Blue) with Ground-Truth filtering.}}
\end{figure}
\FloatBarrier

\newpage
\subsection{SILVR without Filtering}
Visual plans and their executions for SILVR without filtering are illustrated below.

\begin{figure}[ht]
    \centering
    \includegraphics[width=\linewidth]{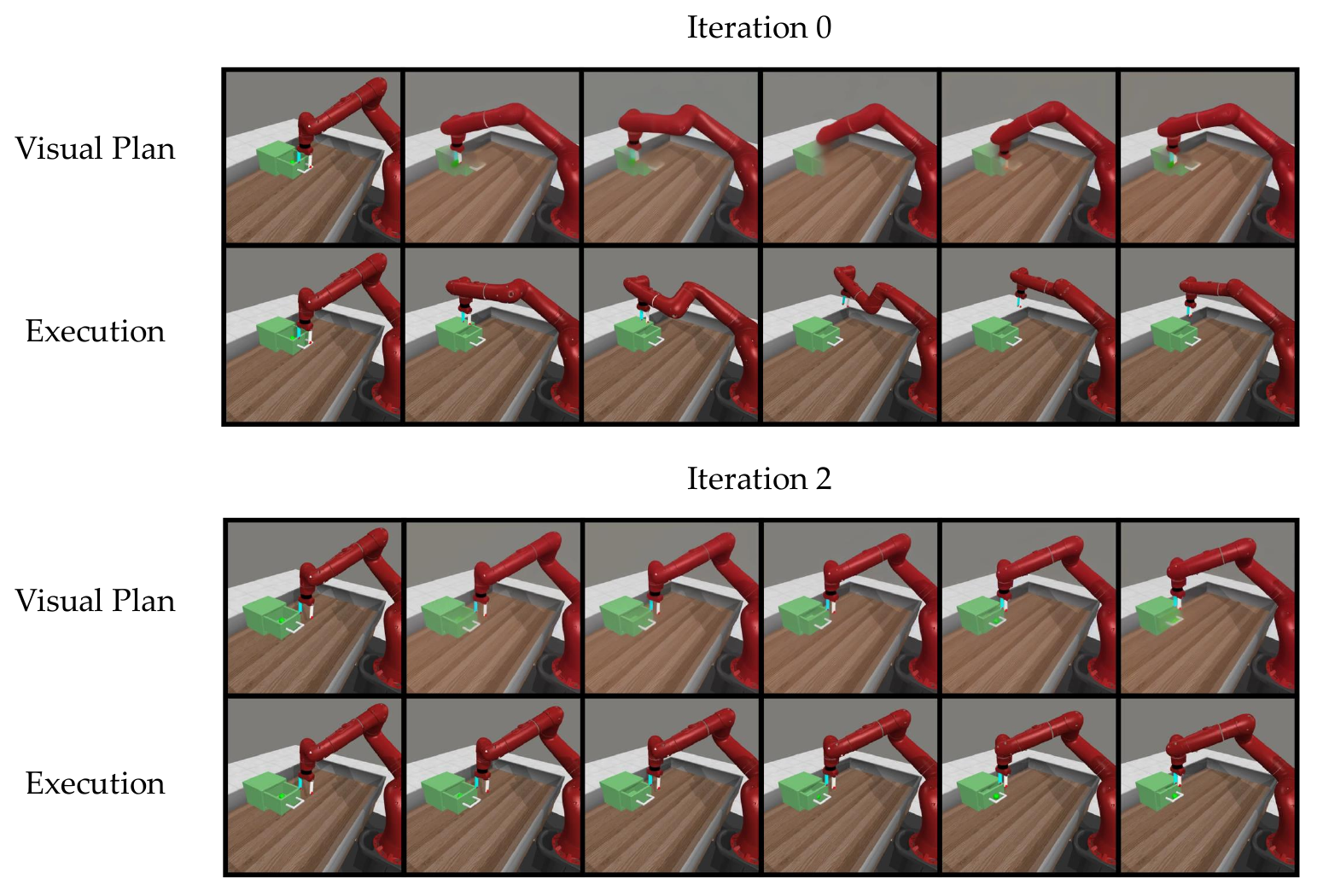}
    \caption{\textbf{SILVR on Drawer Close without filtering.}}
\end{figure}

\begin{figure}[ht]
    \centering
    \includegraphics[width=\linewidth]{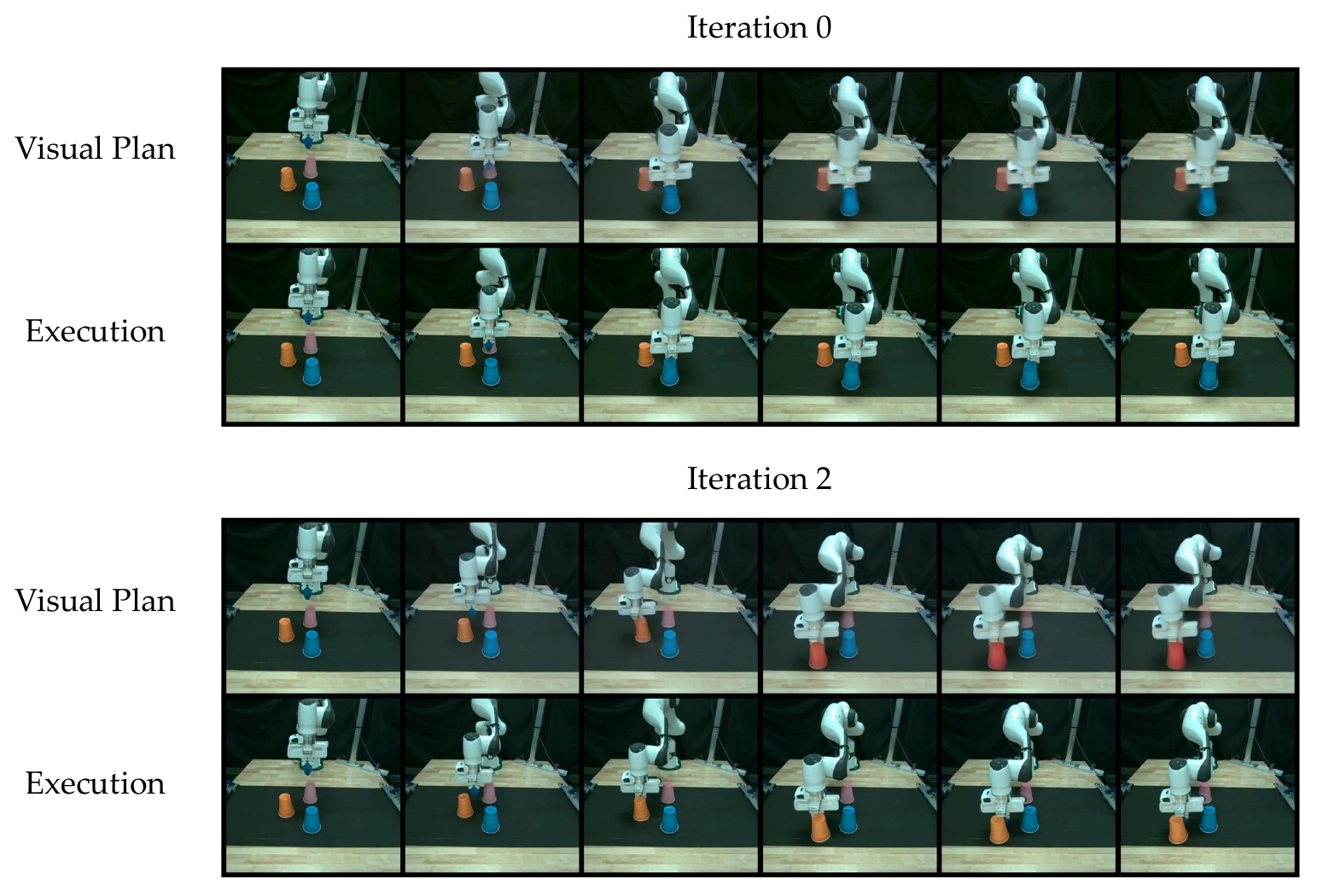}
    \caption{\textbf{SILVR on Orange Cup Pushing (Blue/Pink/Orange) without filtering.}}
\end{figure}

\FloatBarrier
\newpage
\subsection{Corrective Influence of Internet-Scale Video Model Adaptation}

\begin{figure}[ht]
    \centering
    \includegraphics[width=\linewidth]{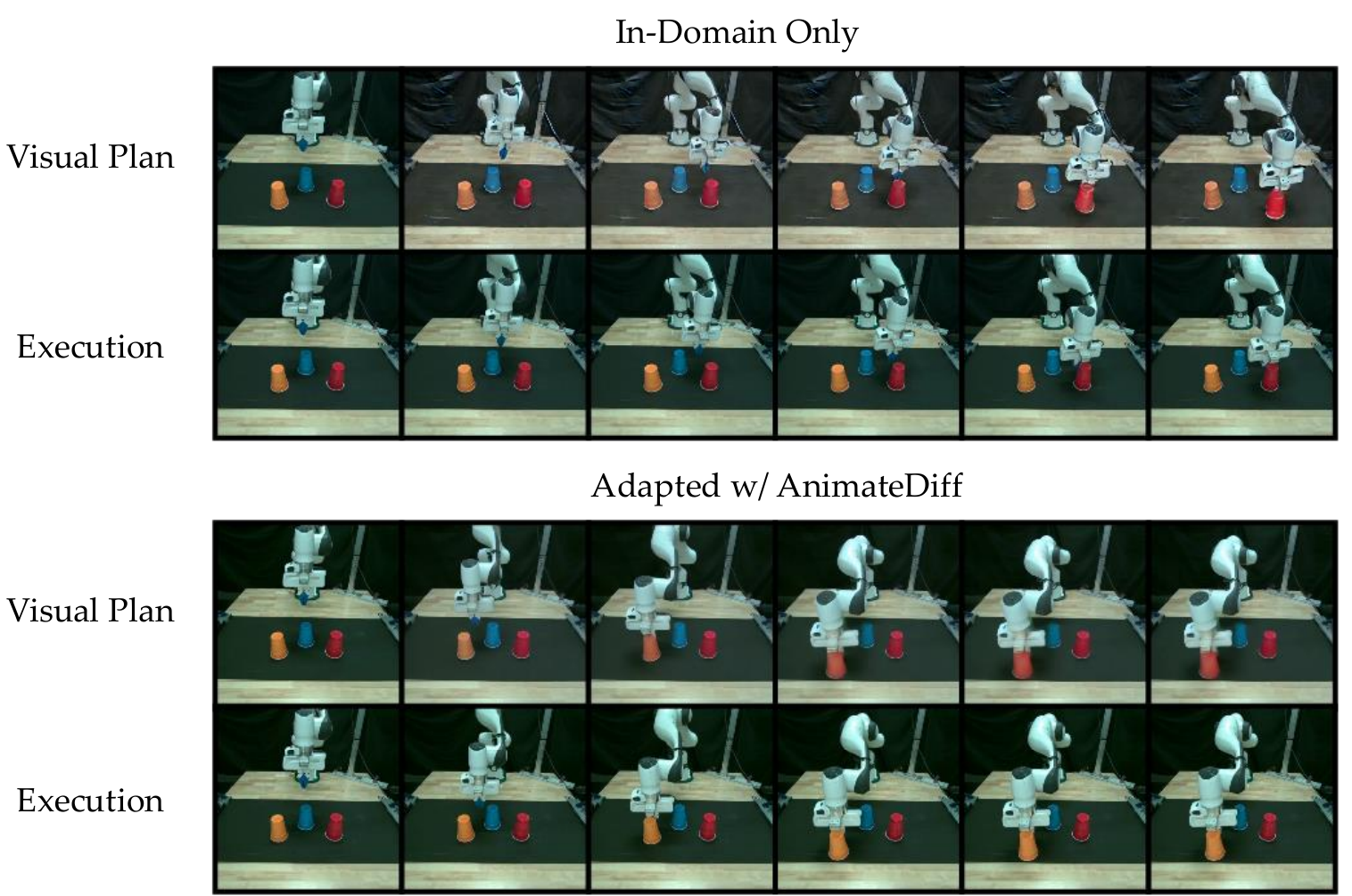}
    \caption{\textbf{Push Orange Cup}}
    \label{fig:ipa_vs_in_domain_orange_cup}
\end{figure}

\begin{figure}[ht]
    \centering
    \includegraphics[width=\linewidth]{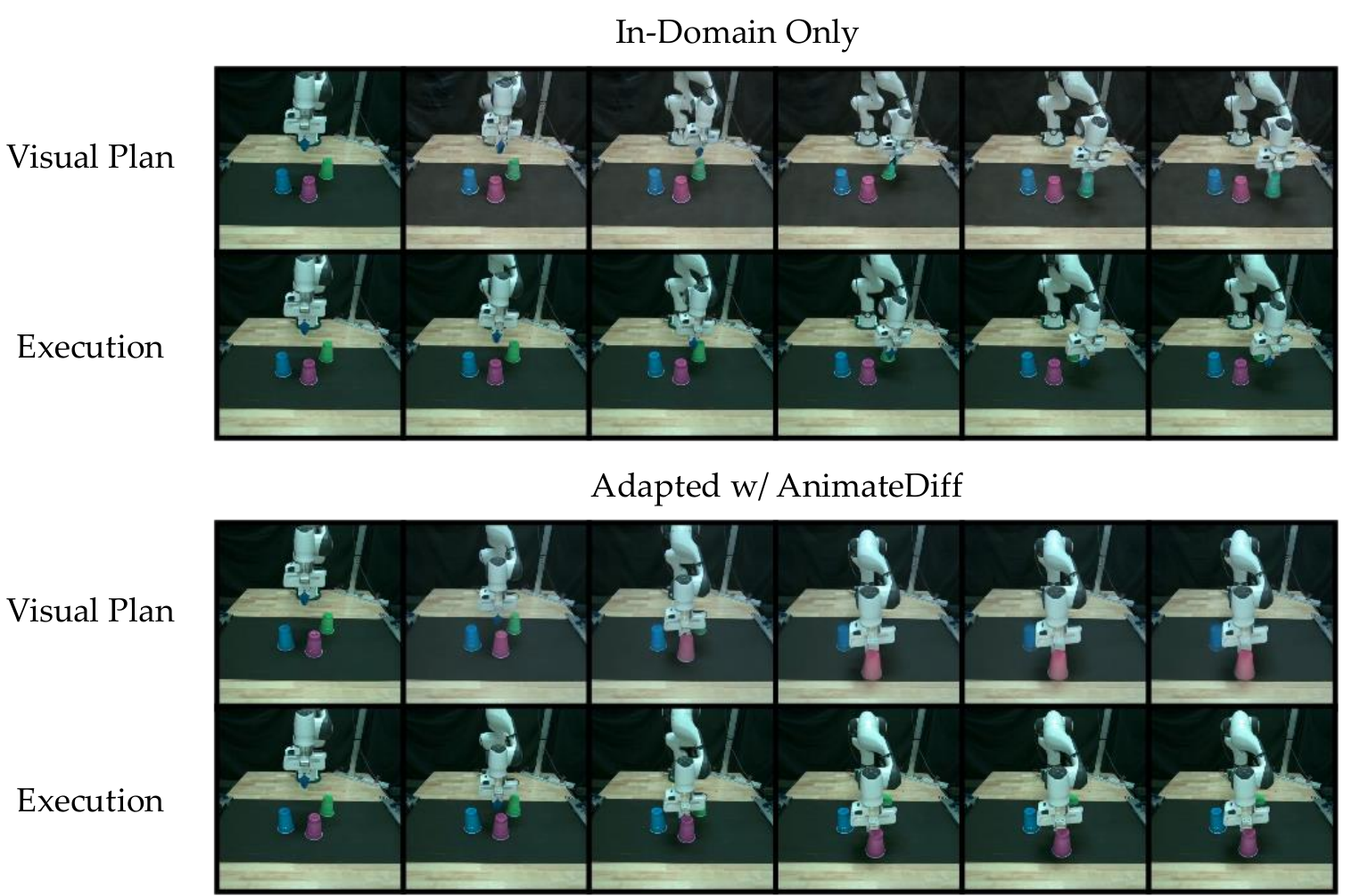}
    \caption{\textbf{Push Purple Cup}}
    \label{fig:ipa_vs_in_domain_purple_cup}
\end{figure}

\begin{figure}[ht]
    \centering
    \includegraphics[width=\linewidth]{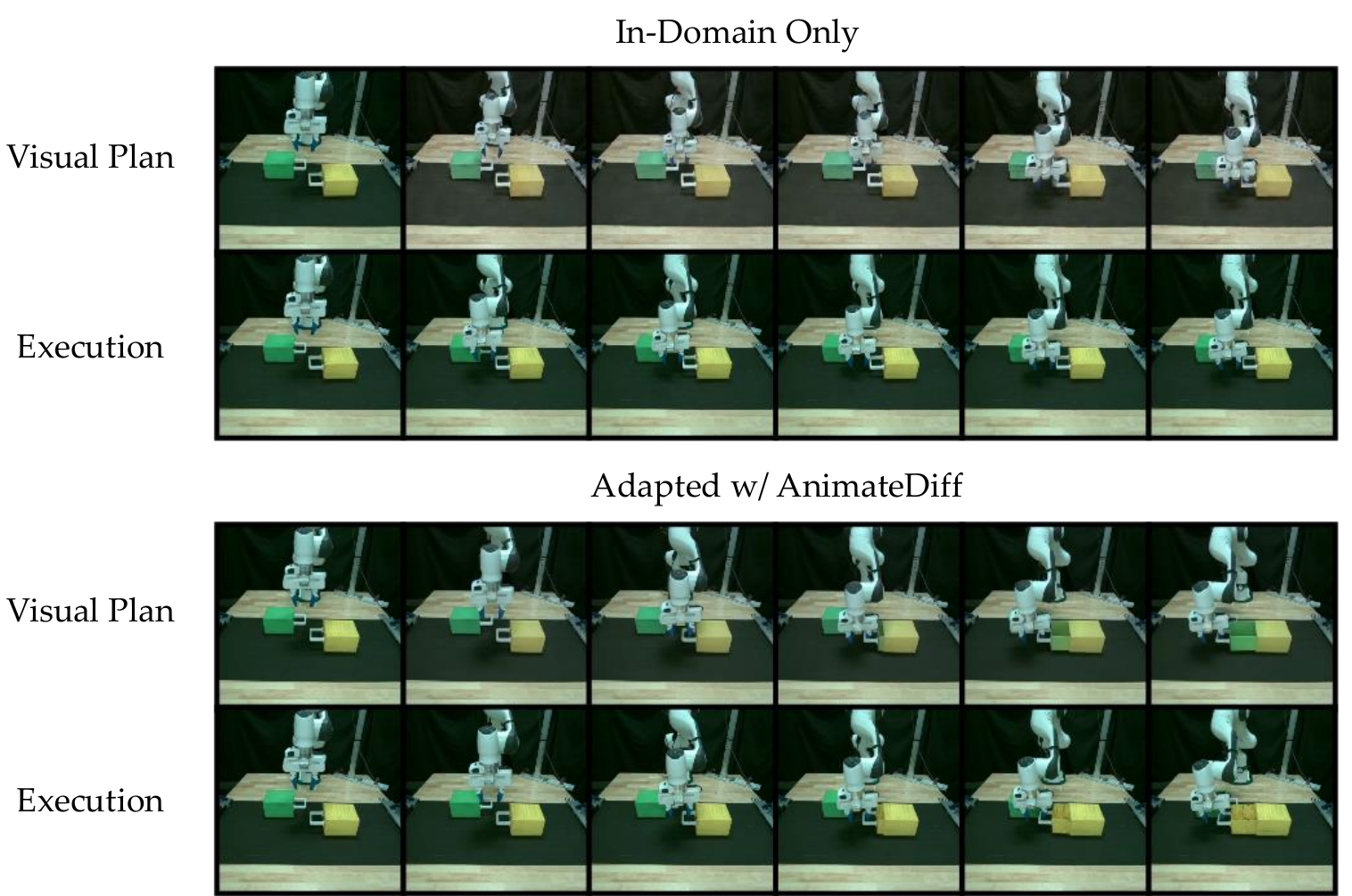}
    \caption{\textbf{Open Yellow Drawer}}
    \label{fig:ipa_vs_in_domain_yellow_drawer}
\end{figure}
\FloatBarrier

\begin{figure}[ht]
    \centering
    \includegraphics[width=\linewidth]{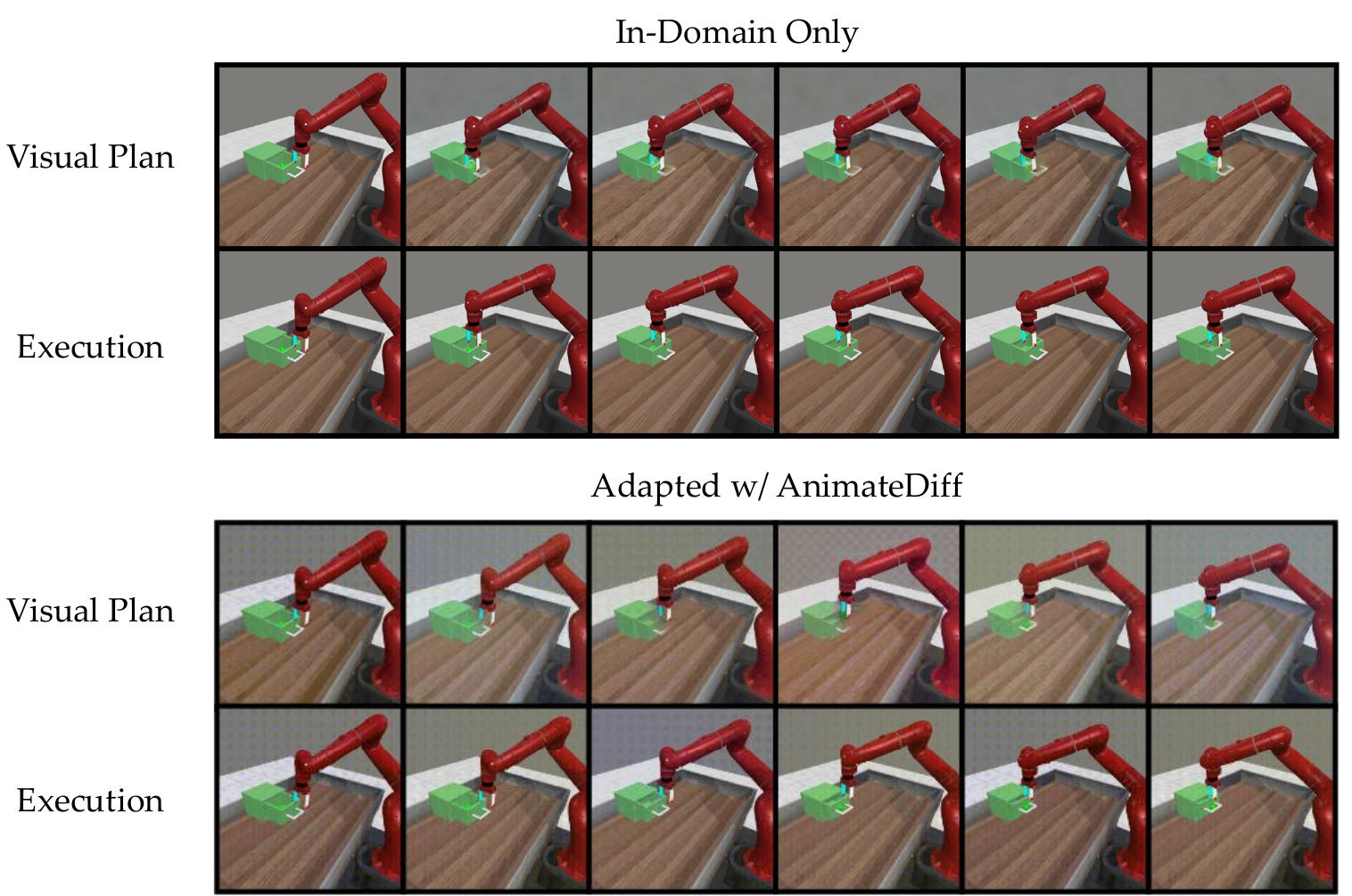}
    \caption{\textbf{Drawer Close}}
    \label{fig:ipa_vs_in_domain_mw_drawer}
\end{figure}
\FloatBarrier

\newpage
\subsection{Long-Horizon Task Evaluation}

Whereas the experiments thus far have demonstrated how SILVR can iteratively improve performance on single tasks, a natural question to consider is how applicable it is to complex tasks.  We note that many long-horizon tasks can be broken down into atomic tasks with shorter horizons, for which SILVR can already demonstrate clear and robust self-improvement behaviors.  We thus investigate how SILVR, by learning individual abilities, naturally facilitates the direct iterative improvement of a sequential long-horizon tasks where success rate depends on the full completion of behaviors in a specific order.

\begin{figure}[ht]
    \centering
    \includegraphics[width=\linewidth]{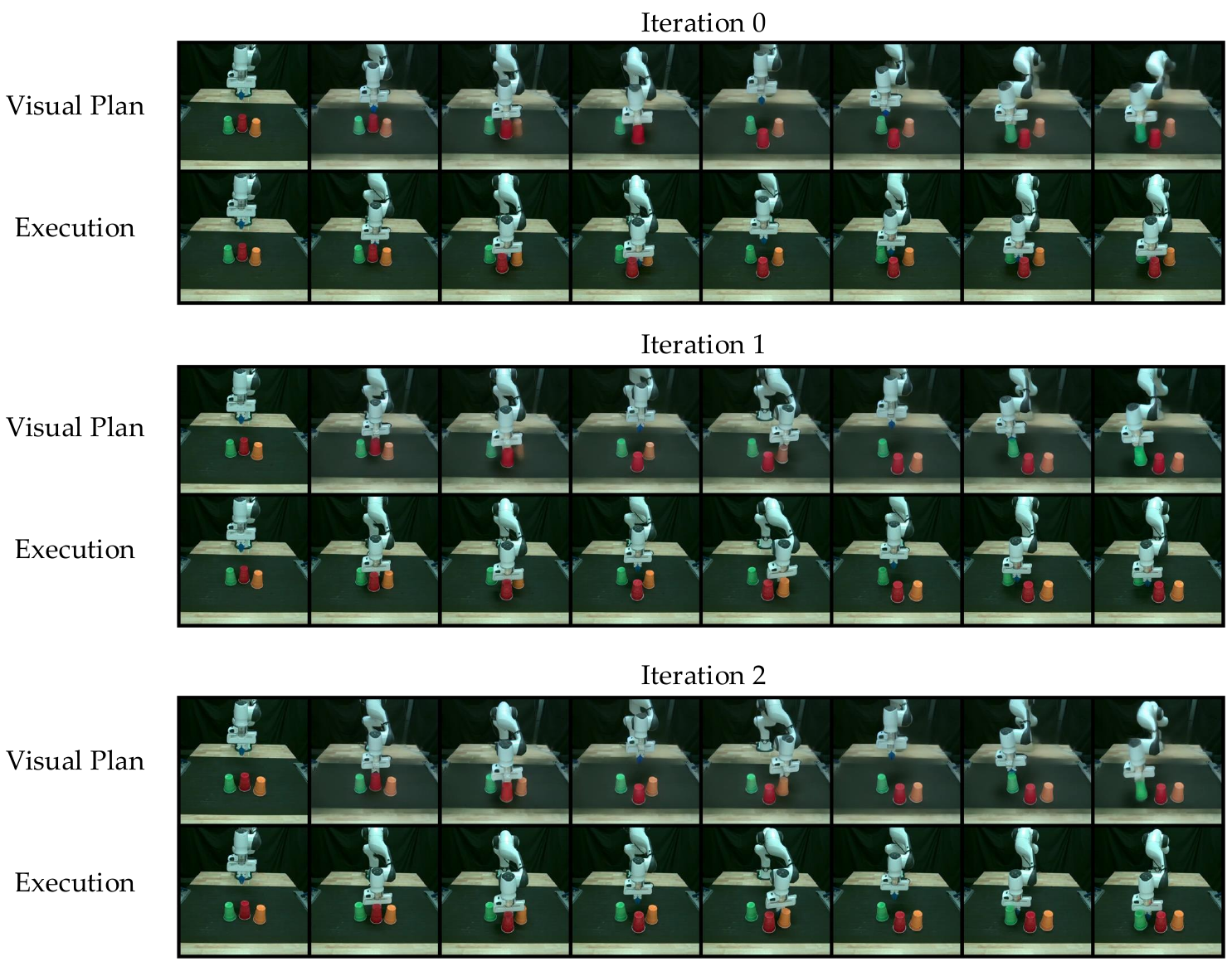}
    \caption{\textbf{Push Cups in the order of Red, Orange, and Green.} The final-iteration visual planner completed the long-horizon task, whereas those from the first two iterations failed.}
    \label{fig:push_rog}
\end{figure}

In particular, we construct an experiment of pushing cups in an order specified by natural language.  For example, in Figure~\ref{fig:push_rog} the prompt is ``Push Cups in the order of Red, Orange, Green", where success is only satisfied if done so in the correct order.  We evaluate SILVR checkpoints automatically using a VLM, implemented as Gemini 2.5, to determine what atomic skill (e.g., pushing the cup of a specific color) to currently execute from the instruction sequence and when to switch to the subsequent one by querying if the visual execution of the current atomic skill was successful.

We find that over successive iterations, despite only applying SILVR to individual atomic tasks, the performance on the overall long-horizon tasks improves.  This suggests that for arbitrary complex long-horizon tasks of interest, SILVR can be combined with an efficient subtask segmentation mechanism to improve individual performance on atomic behaviors and thus improve the final overall performance on long-horizon sequential compositions.

\begin{figure}[ht]
    \centering
    \includegraphics[width=\linewidth]{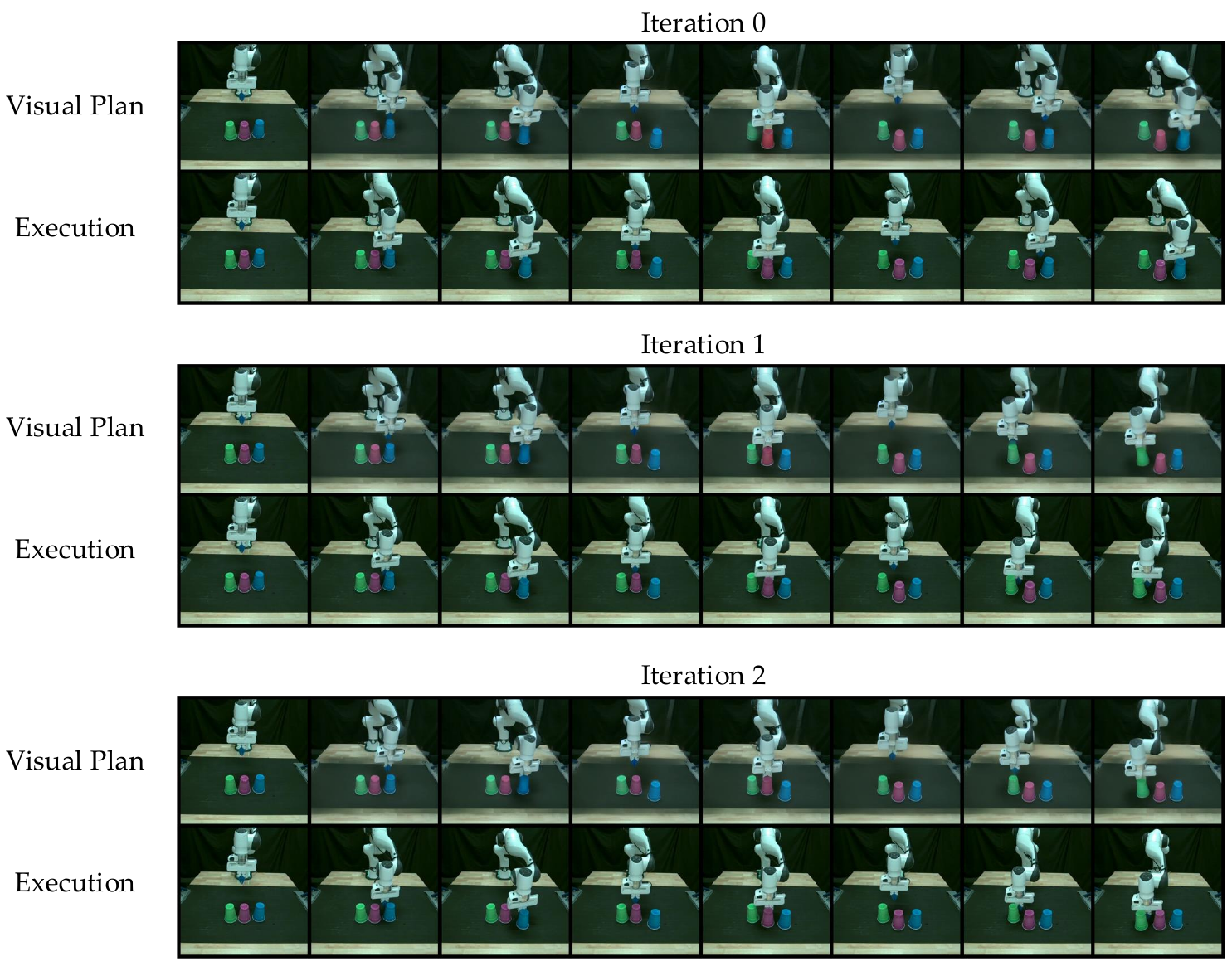}
    \caption{\textbf{Push Cups in the order of Blue, Purple, and Green.} The visual planners of the last two iterations completed the long-horizon task, whereas the one from the first iteration failed.}
    \label{fig:push_bpg}
\end{figure}

\FloatBarrier
\subsection{Visualizing Failure Cases of Final-Iteration Visual Planners}
\begin{figure}[ht]
    \centering
    \includegraphics[width=\linewidth]{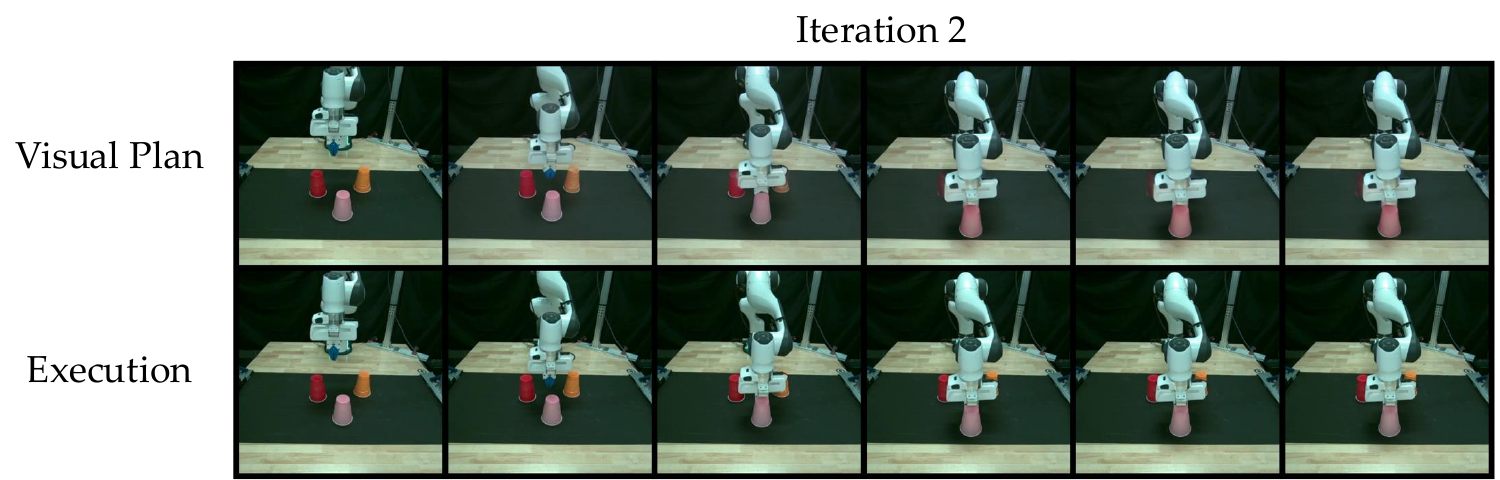}
    \caption{\textbf{Push the Orange Cup (Iteration 2)}}
    \label{fig:failure_mode_final_orange}
\end{figure}

\begin{figure}[ht]
    \centering
    \includegraphics[width=\linewidth]{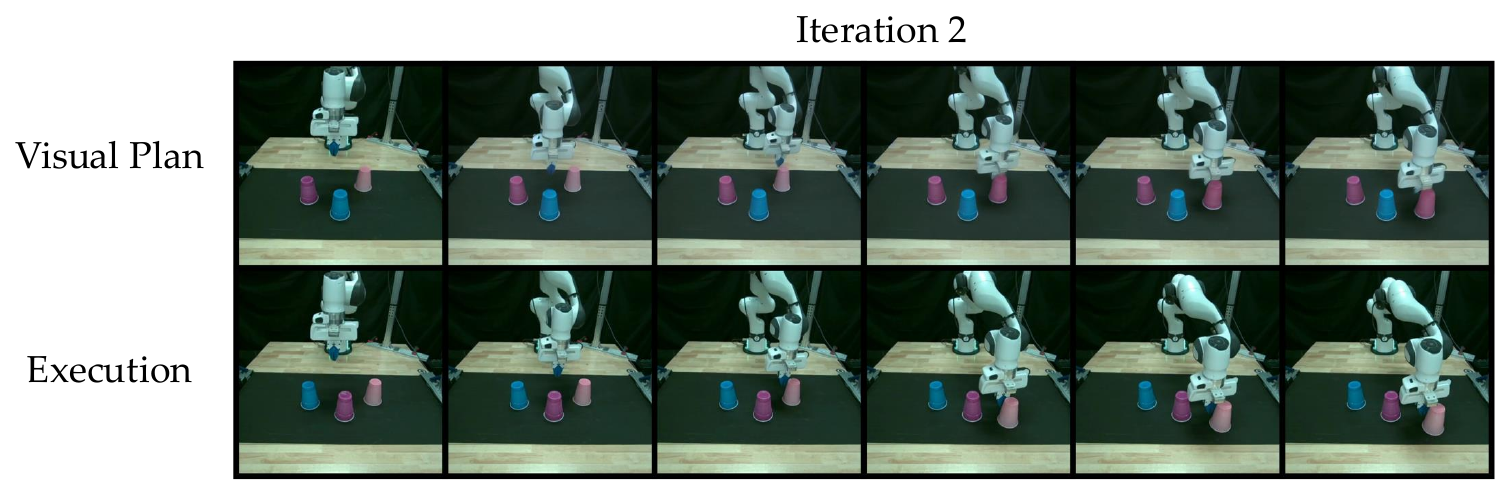}
    \caption{\textbf{Push the Purple Cup (Iteration 2) }}
    \label{fig:failure_mode_final_purple}
\end{figure}

\begin{figure}[ht]
    \centering
    \includegraphics[width=\linewidth]{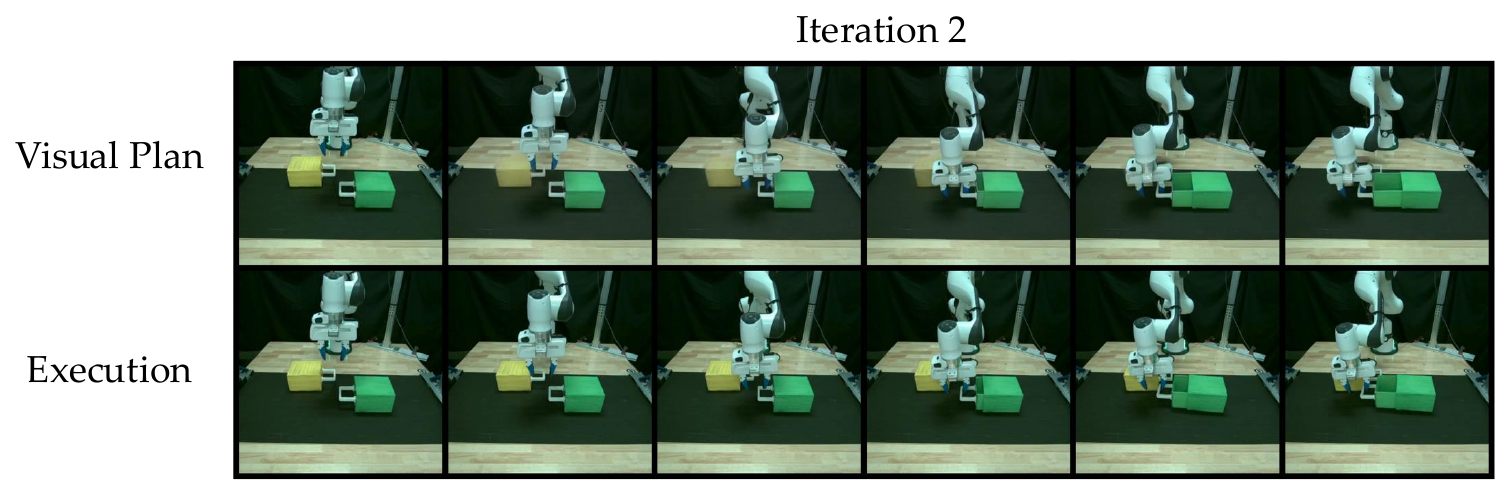}
    \caption{\textbf{Open the Yellow Drawer (Iteration 2)}}
    \label{fig:failure_mode_final_yellow}
\end{figure}

\begin{figure}[ht]
    \centering
    \includegraphics[width=\linewidth]{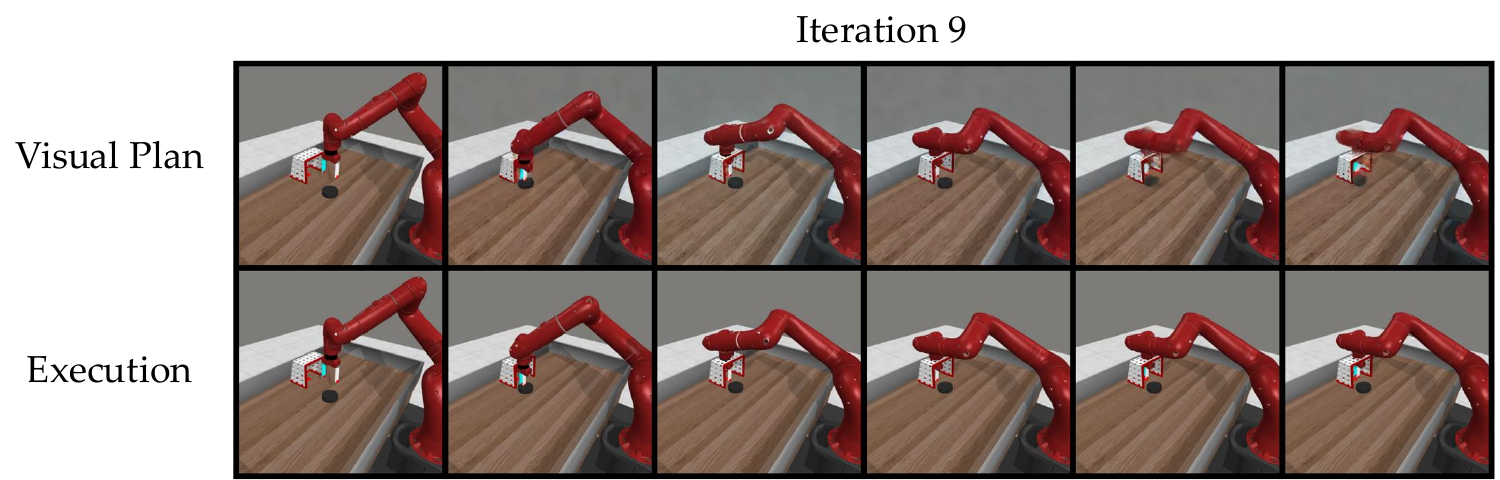}
    \caption{\textbf{Plate Slide (Iteration 9)}}
    \label{fig:failure_mode_final_plate_slide}
\end{figure}

\begin{figure}[ht]
    \centering
    \includegraphics[width=\linewidth]{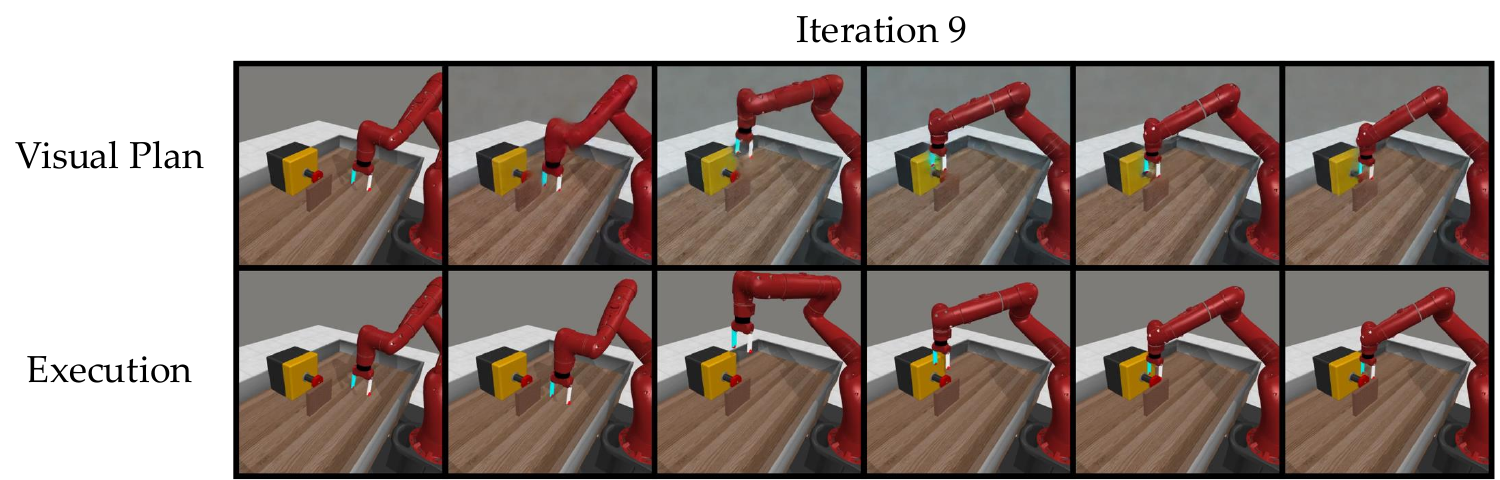}
    \caption{\textbf{Button Press Wall (Iteration 9)}}
    \label{fig:failure_mode_final_button_press}
\end{figure}

Below we visualize the failure cases from the final-iteration visual planners across both real-world and simulation task settings. Most real-world failures arise from semantically incorrect visual plans, in which the robot arm attempts to push the cup (Figure~\ref{fig:failure_mode_final_orange} and~\ref{fig:failure_mode_final_purple}) or open the drawer (Figure~\ref{fig:failure_mode_final_yellow}) of a wrong color, whereas in simulation, we observe a mixture of execution (Figure~\ref{fig:failure_mode_final_plate_slide}) and semantic  (Figure~\ref{fig:failure_mode_final_button_press}) errors, both of which can contribute to unsuccessful task completion.

\FloatBarrier


\begin{thebibliography}{36}
\providecommand{\natexlab}[1]{#1}
\providecommand{\url}[1]{\texttt{#1}}
\expandafter\ifx\csname urlstyle\endcsname\relax
  \providecommand{\doi}[1]{doi: #1}\else
  \providecommand{\doi}{doi: \begingroup \urlstyle{rm}\Url}\fi

\bibitem[Ajay et~al.(2023)Ajay, Han, Du, Li, Gupta, Jaakkola, Tenenbaum, Kaelbling, Srivastava, and Agrawal]{ajaj2023hip}
Anurag Ajay, Seungwook Han, Yilun Du, Shuang Li, Abhi Gupta, Tommi Jaakkola, Josh Tenenbaum, Leslie Kaelbling, Akash Srivastava, and Pulkit Agrawal.
\newblock Compositional foundation models for hierarchical planning.
\newblock In \emph{Conference on Neural Information Processing Systems (NeurIPS)}, 2023.

\bibitem[Ankile et~al.(2025)Ankile, Simeonov, Shenfeld, Torne, and Agrawal]{ankile2025imitation}
Lars Ankile, Anthony Simeonov, Idan Shenfeld, Marcel Torne, and Pulkit Agrawal.
\newblock From imitation to refinement-residual rl for precise assembly.
\newblock In \emph{2025 IEEE International Conference on Robotics and Automation (ICRA)}, pp.\  01--08. IEEE, 2025.

\bibitem[Bain et~al.(2021)Bain, Nagrani, Varol, and Zisserman]{bain2021webvid}
Max Bain, Arsha Nagrani, Gül Varol, and Andrew Zisserman.
\newblock Frozen in time: A joint video and image encoder for end-to-end retrieval.
\newblock In \emph{IEEE International Conference on Computer Vision (ICCV)}, 2021.

\bibitem[Brooks et~al.(2024)Brooks, Peebles, Holmes, DePue, Guo, Jing, Schnurr, Taylor, Luhman, Luhman, Ng, Wang, and Ramesh]{videoworldsimulators2024}
Tim Brooks, Bill Peebles, Connor Holmes, Will DePue, Yufei Guo, Li~Jing, David Schnurr, Joe Taylor, Troy Luhman, Eric Luhman, Clarence Ng, Ricky Wang, and Aditya Ramesh.
\newblock Video generation models as world simulators.
\newblock \emph{OpenAI Blog}, 2024.
\newblock URL \url{https://openai.com/research/video-generation-models-as-world-simulators}.

\bibitem[Bruce et~al.(2024)Bruce, Dennis, Edwards, Parker-Holder, Shi, Hughes, Lai, Mavalankar, Steigerwald, Apps, Aytar, Bechtle, Behbahani, Chan, Heess, Gonzalez, Osindero, Ozair, Reed, Zhang, Zolna, Clune, de~Freitas, Singh, and Rocktaschel]{Bruce2024GenieGI}
Jake Bruce, Michael Dennis, Ashley Edwards, Jack Parker-Holder, Yuge Shi, Edward Hughes, Matthew Lai, Aditi Mavalankar, Richie Steigerwald, Chris Apps, Yusuf Aytar, Sarah Bechtle, Feryal M.~P. Behbahani, Stephanie Chan, Nicolas Manfred~Otto Heess, Lucy Gonzalez, Simon Osindero, Sherjil Ozair, Scott Reed, Jingwei Zhang, Konrad Zolna, Jeff Clune, Nando de~Freitas, Satinder Singh, and Tim Rocktaschel.
\newblock Genie: Generative interactive environments.
\newblock \emph{arXiv preprint arXiv:2402.15391}, 2024.

\bibitem[Chi et~al.(2023)Chi, Feng, Du, Xu, Cousineau, Burchfiel, and Song]{chi2023diffusionpolicy}
Cheng Chi, Siyuan Feng, Yilun Du, Zhenjia Xu, Eric Cousineau, Benjamin Burchfiel, and Shuran Song.
\newblock Diffusion policy: Visuomotor policy learning via action diffusion.
\newblock In \emph{Proceedings of Robotics: Science and Systems (RSS)}, 2023.

\bibitem[Du et~al.(2024{\natexlab{a}})Du, Yang, Dai, Dai, Nachum, Tenenbaum, Schuurmans, and Abbeel]{du2024learning}
Yilun Du, Sherry Yang, Bo~Dai, Hanjun Dai, Ofir Nachum, Josh Tenenbaum, Dale Schuurmans, and Pieter Abbeel.
\newblock Learning universal policies via text-guided video generation.
\newblock \emph{Advances in Neural Information Processing Systems}, 36, 2024{\natexlab{a}}.

\bibitem[Du et~al.(2024{\natexlab{b}})Du, Yang, Florence, Xia, Wahid, brian ichter, Sermanet, Yu, Abbeel, Tenenbaum, Kaelbling, Zeng, and Tompson]{du2024video}
Yilun Du, Sherry Yang, Pete Florence, Fei Xia, Ayzaan Wahid, brian ichter, Pierre Sermanet, Tianhe Yu, Pieter Abbeel, Joshua~B. Tenenbaum, Leslie~Pack Kaelbling, Andy Zeng, and Jonathan Tompson.
\newblock Video language planning.
\newblock In \emph{International Conference on Learning Representations (ICLR)}, 2024{\natexlab{b}}.

\bibitem[Escontrela et~al.(2023)Escontrela, Adeniji, Yan, Jain, Peng, Goldberg, Lee, Hafner, and Abbeel]{alejandro2023viper}
Alejandro Escontrela, Ademi Adeniji, Wilson Yan, Ajay Jain, Xue~Bin Peng, Ken Goldberg, Youngwoon Lee, Danijar Hafner, and Pieter Abbeel.
\newblock Video prediction models as rewards for reinforcement learning.
\newblock In \emph{Conference on Neural Information Processing Systems (NeurIPS)}, 2023.

\bibitem[Guo et~al.(2023)Guo, Yang, Rao, Wang, Qiao, Lin, and Dai]{guo2023animatediff}
Yuwei Guo, Ceyuan Yang, Anyi Rao, Yaohui Wang, Yu~Qiao, Dahua Lin, and Bo~Dai.
\newblock Animatediff: Animate your personalized text-to-image diffusion models without specific tuning.
\newblock \emph{arXiv preprint arXiv:2307.04725}, 2023.

\bibitem[Guo et~al.(2024)Guo, Yang, Rao, Agrawala, Lin, and Dai]{Guo2024ECCV_SparseCtrl_Adding_Sparse}
Yuwei Guo, Ceyuan Yang, Anyi Rao, Maneesh Agrawala, Dahua Lin, and Bo~Dai.
\newblock {SparseCtrl:} adding sparse controls to text-to-video diffusion models.
\newblock In \emph{European Conference on Computer Vision (ECCV)}, 2024.

\bibitem[Huang et~al.(2022)Huang, Gu, Hou, Wu, Wang, Yu, and Han]{Huang2022CONFERENCE_Large_Language_Models}
Jiaxin Huang, S.~Gu, Le~Hou, Yuexin Wu, Xuezhi Wang, Hongkun Yu, and Jiawei Han.
\newblock Large language models can self-improve.
\newblock In \emph{Conference on Empirical Methods in Natural Language Processing}, 2022.

\bibitem[Huang et~al.(2023)Huang, Jiang, Ze, and Xu]{huang2023diffusion}
Tao Huang, Guangqi Jiang, Yanjie Ze, and Huazhe Xu.
\newblock Diffusion reward: Learning rewards via conditional video diffusion.
\newblock \emph{arXiv preprint arXiv:2312.14134}, 2023.

\bibitem[Ko et~al.(2024)Ko, Mao, Du, Sun, and Tenenbaum]{ko2024avdc}
Po-Chen Ko, Jiayuan Mao, Yilun Du, Shao-Hua Sun, and Joshua~B. Tenenbaum.
\newblock Learning to act from actionless videos through dense correspondences.
\newblock In \emph{International Conference on Learning Representations (ICLR)}, 2024.

\bibitem[Liang et~al.(2024)Liang, Liu, Ozguroglu, Sudhakar, Dave, Tokmakov, Song, and Vondrick]{liang2024dreamitate}
Junbang Liang, Ruoshi Liu, Ege Ozguroglu, Sruthi Sudhakar, Achal Dave, Pavel Tokmakov, Shuran Song, and Carl Vondrick.
\newblock Dreamitate: Real-world visuomotor policy learning via video generation.
\newblock \emph{arXiv preprint arXiv:2406.16862}, 2024.

\bibitem[Luo et~al.(2024)Luo, He, Zeng, and Sun]{luo2024text}
Calvin Luo, Mandy He, Zilai Zeng, and Chen Sun.
\newblock Text-aware diffusion for policy learning.
\newblock In \emph{Advances in Neural Information Processing Systems}, volume~37, 2024.

\bibitem[Luo et~al.(2025)Luo, Zeng, Du, and Sun]{luo2025solving}
Calvin Luo, Zilai Zeng, Yilun Du, and Chen Sun.
\newblock Solving new tasks by adapting internet video knowledge.
\newblock In \emph{The Thirteenth International Conference on Learning Representations}, 2025.
\newblock URL \url{https://openreview.net/forum?id=p01BR4njlY}.

\bibitem[Majumdar et~al.(2023)Majumdar, Yadav, Arnaud, Ma, Chen, Silwal, Jain, Berges, Wu, Vakil, Abbeel, Malik, Batra, Lin, Maksymets, Rajeswaran, and Meier]{majumdar2023vc1}
Arjun Majumdar, Karmesh Yadav, Sergio Arnaud, Jason Ma, Claire Chen, Sneha Silwal, Aryan Jain, Vincent-Pierre Berges, Tingfan Wu, Jay Vakil, Pieter Abbeel, Jitendra Malik, Dhruv Batra, Yixin Lin, Oleksandr Maksymets, Aravind Rajeswaran, and Franziska Meier.
\newblock Where are we in the search for an artificial visual cortex for embodied intelligence?
\newblock In \emph{Conference on Neural Information Processing Systems (NeurIPS)}, 2023.

\bibitem[McCarthy et~al.(2024)McCarthy, Tan, Schmidt, Acero, Herr, Du, Thuruthel, and Li]{McCarthy2024TowardsGR}
Robert McCarthy, Daniel~C.H. Tan, Dominik Schmidt, Fernando Acero, Nathan Herr, Yilun Du, Thomas~George Thuruthel, and Zhibin Li.
\newblock Towards generalist robot learning from internet video: A survey.
\newblock \emph{arXiv preprint arXiv:2404.19664}, 2024.

\bibitem[Patel et~al.(2024)Patel, Hofmarcher, Leoveanu-Condrei, Dinu, Callison-Burch, and Hochreiter]{Patel2024ARXIV_Large_Language_Models}
Ajay Patel, Markus Hofmarcher, Claudiu Leoveanu-Condrei, Marius-Constantin Dinu, Chris Callison-Burch, and Sepp Hochreiter.
\newblock Large language models can self-improve at web agent tasks.
\newblock \emph{arXiv}, 2405.20309, 2024.

\bibitem[Ren et~al.(2024)Ren, Lidard, Ankile, Simeonov, Agrawal, Majumdar, Burchfiel, Dai, and Simchowitz]{ren2024diffusion}
Allen~Z Ren, Justin Lidard, Lars~L Ankile, Anthony Simeonov, Pulkit Agrawal, Anirudha Majumdar, Benjamin Burchfiel, Hongkai Dai, and Max Simchowitz.
\newblock Diffusion policy policy optimization.
\newblock \emph{arXiv preprint arXiv:2409.00588}, 2024.

\bibitem[Song et~al.(2021)Song, Meng, and Ermon]{song2021ddim}
Jiaming Song, Chenlin Meng, and Stefano Ermon.
\newblock Denoising diffusion implicit models.
\newblock In \emph{International Conference on Learning Representations (ICLR)}, 2021.

\bibitem[Soni et~al.(2024)Soni, Venkataraman, Chandra, Fischmeister, Liang, Dai, and Yang]{Soni2024VideoAgentSV}
Achint Soni, Sreyas Venkataraman, Abhranil Chandra, Sebastian Fischmeister, Percy Liang, Bo~Dai, and Sherry Yang.
\newblock Videoagent: Self-improving video generation.
\newblock \emph{arXiv preprint arXiv:2410.10076}, 2024.

\bibitem[Tian et~al.(2024)Tian, Peng, Song, Jin, Yu, Han, Mi, and Yu]{Tian2024NEURIPS_Toward_Self_Improvement}
Ye~Tian, Baolin Peng, Linfeng Song, Lifeng Jin, Dian Yu, Lei Han, Haitao Mi, and Dong Yu.
\newblock Toward self-improvement of {LLMs} via imagination, searching, and criticizing.
\newblock In \emph{Conference on Neural Information Processing Systems (NeurIPS)}, 2024.

\bibitem[Valevski et~al.(2024)Valevski, Leviathan, Arar, and Fruchter]{Valevski2024DiffusionMA}
Dani Valevski, Yaniv Leviathan, Moab Arar, and Shlomi Fruchter.
\newblock Diffusion models are real-time game engines.
\newblock \emph{arXiv preprint arXiv:2408.14837}, 2024.

\bibitem[Veo-Team et~al.(2024)Veo-Team, :, Gupta, Razavi, Toor, Gupta, Erhan, Shaw, Lau, Belletti, Barth-Maron, Shaw, Erdogan, Sidahmed, Nandwani, Moraldo, Kim, Blok, Donahue, Lezama, Mathewson, David, Lorrain, van Zee, Narasimhan, Wang, Babaeizadeh, Papalampidi, Pezzotti, Jha, Barnes, Kindermans, Hornung, Villegas, Poplin, Zaiem, Dieleman, Ebrahimi, Wisdom, Zhang, Fruchter, Nørly, Hua, Yan, Du, and Chen]{veo2}
Veo-Team, :, Agrim Gupta, Ali Razavi, Andeep Toor, Ankush Gupta, Dumitru Erhan, Eleni Shaw, Eric Lau, Frank Belletti, Gabe Barth-Maron, Gregory Shaw, Hakan Erdogan, Hakim Sidahmed, Henna Nandwani, Hernan Moraldo, Hyunjik Kim, Irina Blok, Jeff Donahue, José Lezama, Kory Mathewson, Kurtis David, Matthieu~Kim Lorrain, Marc van Zee, Medhini Narasimhan, Miaosen Wang, Mohammad Babaeizadeh, Nelly Papalampidi, Nick Pezzotti, Nilpa Jha, Parker Barnes, Pieter-Jan Kindermans, Rachel Hornung, Ruben Villegas, Ryan Poplin, Salah Zaiem, Sander Dieleman, Sayna Ebrahimi, Scott Wisdom, Serena Zhang, Shlomi Fruchter, Signe Nørly, Weizhe Hua, Xinchen Yan, Yuqing Du, and Yutian Chen.
\newblock Veo 2.
\newblock 2024.
\newblock URL \url{https://deepmind.google/technologies/veo/veo-2/}.

\bibitem[Wagenmaker et~al.(2025)Wagenmaker, Nakamoto, Zhang, Park, Yagoub, Nagabandi, Gupta, and Levine]{wagenmaker2025steering}
Andrew Wagenmaker, Mitsuhiko Nakamoto, Yunchu Zhang, Seohong Park, Waleed Yagoub, Anusha Nagabandi, Abhishek Gupta, and Sergey Levine.
\newblock Steering your diffusion policy with latent space reinforcement learning.
\newblock \emph{arXiv preprint arXiv:2506.15799}, 2025.

\bibitem[Wang et~al.(2025)Wang, Ai, Wen, Mao, Xie, Chen, Yu, Zhao, Yang, Zeng, Wang, Zhang, Zhou, Wang, Chen, Zhu, Zhao, Yan, Huang, Feng, Zhang, Li, Wu, Chu, Feng, Zhang, Sun, Fang, Wang, Gui, Weng, Shen, Lin, Wang, Wang, Zhou, Wang, Shen, Yu, Shi, Huang, Xu, Kou, Lv, Li, Liu, Wang, Zhang, Huang, Li, Wu, Liu, Pan, Zheng, Hong, Shi, Feng, Jiang, Han, Wu, and Liu]{wan2025}
Ang Wang, Baole Ai, Bin Wen, Chaojie Mao, Chen-Wei Xie, Di~Chen, Feiwu Yu, Haiming Zhao, Jianxiao Yang, Jianyuan Zeng, Jiayu Wang, Jingfeng Zhang, Jingren Zhou, Jinkai Wang, Jixuan Chen, Kai Zhu, Kang Zhao, Keyu Yan, Lianghua Huang, Mengyang Feng, Ningyi Zhang, Pandeng Li, Pingyu Wu, Ruihang Chu, Ruili Feng, Shiwei Zhang, Siyang Sun, Tao Fang, Tianxing Wang, Tianyi Gui, Tingyu Weng, Tong Shen, Wei Lin, Wei Wang, Wei Wang, Wenmeng Zhou, Wente Wang, Wenting Shen, Wenyuan Yu, Xianzhong Shi, Xiaoming Huang, Xin Xu, Yan Kou, Yangyu Lv, Yifei Li, Yijing Liu, Yiming Wang, Yingya Zhang, Yitong Huang, Yong Li, You Wu, Yu~Liu, Yulin Pan, Yun Zheng, Yuntao Hong, Yupeng Shi, Yutong Feng, Zeyinzi Jiang, Zhen Han, Zhi-Fan Wu, and Ziyu Liu.
\newblock Wan: Open and advanced large-scale video generative models.
\newblock \emph{arXiv preprint arXiv:2503.20314}, 2025.

\bibitem[Yang et~al.(2023{\natexlab{a}})Yang, Du, Dai, Schuurmans, Tenenbaum, and Abbeel]{yang2023probabilistic}
Mengjiao Yang, Yilun Du, Bo~Dai, Dale Schuurmans, Joshua~B Tenenbaum, and Pieter Abbeel.
\newblock Probabilistic adaptation of text-to-video models.
\newblock \emph{arXiv preprint arXiv:2306.01872}, 2023{\natexlab{a}}.

\bibitem[Yang et~al.(2023{\natexlab{b}})Yang, Du, Ghasemipour, Tompson, Schuurmans, and Abbeel]{yang2023learning}
Mengjiao Yang, Yilun Du, Kamyar Ghasemipour, Jonathan Tompson, Dale Schuurmans, and Pieter Abbeel.
\newblock Learning interactive real-world simulators.
\newblock \emph{arXiv preprint arXiv:2310.06114}, 2023{\natexlab{b}}.

\bibitem[Yang et~al.(2024{\natexlab{a}})Yang, Walker, Parker-Holder, Du, Bruce, Barreto, Abbeel, and Schuurmans]{yang2024video}
Sherry Yang, Jacob Walker, Jack Parker-Holder, Yilun Du, Jake Bruce, Andre Barreto, Pieter Abbeel, and Dale Schuurmans.
\newblock Video as the new language for real-world decision making.
\newblock \emph{arXiv preprint arXiv:2402.17139}, 2024{\natexlab{a}}.

\bibitem[Yang et~al.(2024{\natexlab{b}})Yang, Teng, Zheng, Ding, Huang, Xu, Yang, Hong, Zhang, Feng, et~al.]{yang2024cogvideox}
Zhuoyi Yang, Jiayan Teng, Wendi Zheng, Ming Ding, Shiyu Huang, Jiazheng Xu, Yuanming Yang, Wenyi Hong, Xiaohan Zhang, Guanyu Feng, et~al.
\newblock Cogvideox: Text-to-video diffusion models with an expert transformer.
\newblock \emph{arXiv preprint arXiv:2408.06072}, 2024{\natexlab{b}}.

\bibitem[Yu et~al.(2020)Yu, Quillen, He, Julian, Hausman, Finn, and Levine]{yu2020meta}
Tianhe Yu, Deirdre Quillen, Zhanpeng He, Ryan Julian, Karol Hausman, Chelsea Finn, and Sergey Levine.
\newblock Meta-world: A benchmark and evaluation for multi-task and meta reinforcement learning.
\newblock In \emph{Conference on robot learning}, pp.\  1094--1100. PMLR, 2020.

\bibitem[Yu et~al.(2024)Yu, Peng, Galley, Gao, and Yu]{yu-etal-2024-teaching}
Xiao Yu, Baolin Peng, Michel Galley, Jianfeng Gao, and Zhou Yu.
\newblock Teaching language models to self-improve through interactive demonstrations.
\newblock In Kevin Duh, Helena Gomez, and Steven Bethard (eds.), \emph{Proceedings of the 2024 Conference of the North American Chapter of the Association for Computational Linguistics: Human Language Technologies (Volume 1: Long Papers)}, pp.\  5127--5149, Mexico City, Mexico, June 2024. Association for Computational Linguistics.
\newblock \doi{10.18653/v1/2024.naacl-long.287}.
\newblock URL \url{https://aclanthology.org/2024.naacl-long.287/}.

\bibitem[Yuan et~al.(2024)Yuan, Pang, Cho, Li, Sukhbaatar, Xu, and Weston]{Yuan2024ICML_Self_Rewarding_Language}
Weizhe Yuan, Richard~Yuanzhe Pang, Kyunghyun Cho, Xian Li, Sainbayar Sukhbaatar, Jing Xu, and Jason~E Weston.
\newblock Self-rewarding language models.
\newblock In \emph{International Conference on Machine Learning (ICML)}, 2024.

\bibitem[Zhou et~al.(2024)Zhou, Du, Chen, Li, Yeung, and Gan]{Zhou2024RoboDreamerLC}
Siyuan Zhou, Yilun Du, Jiaben Chen, Yandong Li, D.~Y. Yeung, and Chuang Gan.
\newblock Robodreamer: Learning compositional world models for robot imagination.
\newblock \emph{arXiv preprint arXiv:2404.12377}, 2024.

\end{thebibliography}
\end{document}